%% file: main.tex
\documentclass{article}

\PassOptionsToPackage{numbers, compress}{natbib}

\usepackage[preprint]{neurips_2025}

\usepackage[utf8]{inputenc} 
\usepackage[T1]{fontenc}    
\usepackage[colorlinks,linkcolor=black,anchorcolor=black,citecolor=black]{hyperref}    
\usepackage{url}            
\usepackage{booktabs}       
\usepackage{amsfonts}       
\usepackage{nicefrac}       
\usepackage{microtype}      
\usepackage{xcolor}         
\usepackage{graphicx}
\usepackage{subfigure}
\usepackage{hyperref}

\usepackage{amsmath}
\usepackage{amssymb}
\usepackage{mathtools}
\usepackage{amsthm}

\usepackage[capitalize,noabbrev]{cleveref}
\usepackage{here}
\usepackage{color}
\usepackage{xspace}
\newcommand{\name}[0]{DAM-GT\xspace} 

\makeatletter
\DeclareRobustCommand\onedot{\futurelet\@let@token\@onedot}
\def\@onedot{\ifx\@let@token.\else.\null\fi\xspace}

\def\ie{\emph{i.e}\onedot, }

\makeatother

\theoremstyle{plain}

\theoremstyle{definition}

\theoremstyle{remark}

\usepackage[textsize=tiny]{todonotes}

\title{DAM-GT: Dual Positional Encoding-Based Attention Masking Graph Transformer for Node Classification}

\author{
  Chenyang Li$^{1,2}$\thanks{The first two authors contribute equally.}~, 
  Jinsong Chen$^{1,2}$\footnotemark[1]~,
  John E. Hopcroft$^{2,3}$, Kun He$^{1,2}\thanks{Corresponding author.}$ \\
  $^1$School of Computer Science and Technology, 
  Huazhong University of Science and Technology\\
  $^2$Hopcroft Center on Computing Science,
  Huazhong University of Science and Technology\\
  $^3$Department of Computer Science, Cornell University \\
  \texttt{ 
  \{chenyangli,chenjinsong\}@hust.edu.cn,}\\
  \texttt{
  jeh@cs.cornell.edu,
  brooklet60@hust.edu.cn} 
}

\begin{document}

\maketitle

\begin{abstract}
  Neighborhood-aware tokenized graph Transformers have recently shown great potential for node classification tasks. Despite their effectiveness,  our in-depth analysis of neighborhood tokens reveals two critical limitations in the existing paradigm. 
  First, current  neighborhood token generation methods fail to adequately capture attribute correlations within a neighborhood. Second, the conventional self-attention mechanism suffers from attention diversion when processing neighborhood tokens, where high-hop neighborhoods receive disproportionate focus, severely disrupting information interactions between the target node and its neighborhood tokens.
  To address these challenges, we propose DAM-GT, Dual positional encoding-based Attention Masking graph Transformer. 
  DAM-GT introduces a novel dual positional encoding scheme that incorporates attribute-aware encoding via an attribute clustering strategy, effectively preserving node correlations in both topological and attribute spaces. In addition, DAM-GT formulates a new attention mechanism with a simple yet effective masking strategy to guide interactions between target nodes and their neighborhood tokens, overcoming the issue of attention diversion. Extensive experiments on various graphs with different homophily levels as well as different scales demonstrate that DAM-GT consistently outperforms state-of-the-art methods in node classification tasks.
\end{abstract}

\section{Introduction}
\label{introduction}
\input{body/01intro}

\section{Preliminary}
\label{preliminary}
\input{body/03preliminary}

\section{Analysis on Attention Scores}
\label{DA}
\input{body/04analysis}

\begin{figure*}[t]
    \centering    
    \includegraphics[width=0.9\linewidth]{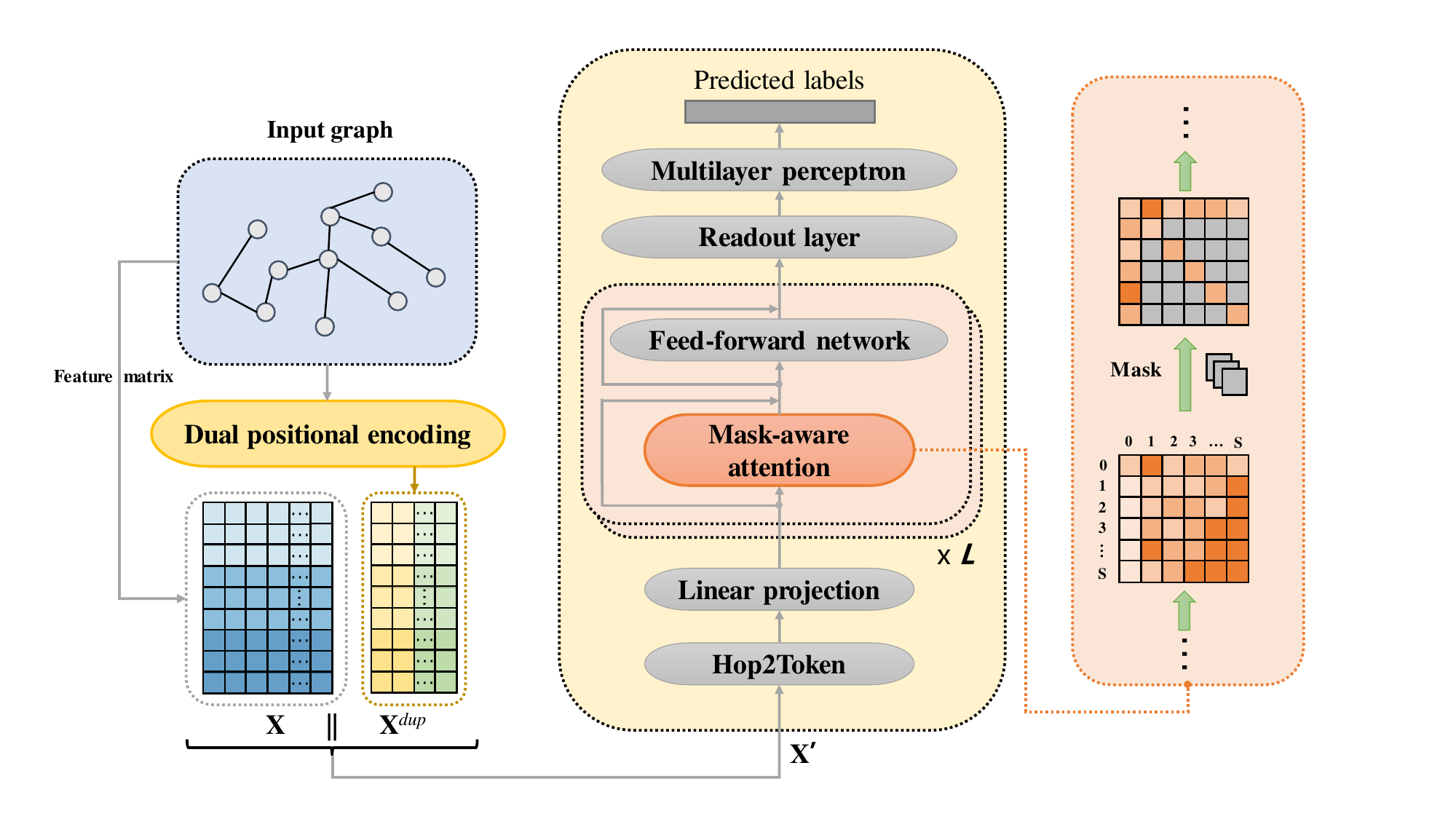}
    \vspace{-1em}
    \caption{The overall view of our \name. 
    Given the input graph, \name first utilizes dual positional encoding to enhance original node features. Then the Hop2Token module is employed to generate neighborhood-aware token sequences as input for the Transformer-based backbone, where \name develops mask-aware self-attention mechanism to learn node representations. Finally, \name adopts a readout layer and a Multilayer perceptron for final label prediction.
    }
    \vspace{-0.5em}
    \label{fig:backbone}
\end{figure*}

\section{Methodology}
\label{methodology}
\input{body/05method}

\section{Experiments} 
\label{exp}

\input{body/06Exp}

\section{Conclusion} 
\label{conclusion}
\input{body/07conclusion}

\bibliographystyle{neurips_2025}
\bibliography{reference}

\newpage
\appendix
\input{body/08APP}

\end{document}

%% file: body/01intro.tex
Node classification is a fundamental task in graph data mining, aimed at predicting labels of unknown nodes within a graph. 
There are numerous related applications, such as fraud detection~\cite{Fraud_Detection_1} and recommendation system~\cite{rs1,rs2}, which have garnered significant interest in the literature.

Graph Neural Networks (GNNs)~\cite{appnp,gprgnn} are popular solutions for this task.
Based on the message-passing mechanism~\cite{message} that aggregates information of immediate neighborhoods, GNNs can simultaneously utilize node attribute information and graph structure information to learn rich node representations, showing impressive performance in node classification.
While effective, GNN-based methods suffer from the inherent limitations of the message-passing mechanism, such as over-smoothing~\cite{oversmoothing} and over-squashing~\cite{oversq}, restricting their ability to capture deep graph structural information.

Recently, graph Transformers~\cite{gps,graphormer,nagformer} have emerged as a powerful alternative to GNNs for graph representation learning.
By incorporating the self-attention mechanism of Transformers, graph Transformers can effectively capture the relations between nodes in the whole graph, naturally breaking through the limitation of message-passing mechanism~\cite{graphormer}.
Nevertheless, due to the quadratic computational complexity of the self-attention mechanism, scalability  is the greatest challenge in generalizing graph Transformers on the node classification task. 

To address the scalability issue, tokenized graph Transformers~\cite{nagformer,ansgt,vcr,polyformer} have attracted increasing attention in recent years.
These models transform the input graph into a series of independent token sequences as the model input, allowing them to utilize mini-batch technique and effectively control the training cost.
Generally speaking, existing tokenized graph Transformers mainly generate two types of tokens, \ie node-wise tokens and neighborhood-wise tokens, to construct the input token sequences. 
In contrast to node-wise tokens yielded by diverse sampling strategies, neighborhood-wise tokens generated through the propagation operation are not only efficiently acquirable but also capable of effectively retaining local neighborhood information, thus showing promising performance in node classification.
Therefore, this paper focuses on neighborhood-aware tokenized graph Transformers. 
A comprehensive review of recent studies is provided in \autoref{app:rw}.

The central idea behind neighborhood-aware tokenized graph Transformers is to leverage the Transformer architecture to learn node representations from multi-hop neighborhood tokens~\cite{nagformer,vcr}. 
Nevertheless, in this paper, we have identified the following two key limitations in current approaches, which provide opportunities for further enhancing model performance in node classification.

\textbf{(1) Ignoring attribute correlations within a neighborhood.}
Existing methods typically propagate node features with topology-aware positional encoding to generate neighborhood-wise tokens. However, they often overlook the semantic attribute correlations between nodes in the attribute space.
The inability to efficiently preserve these attribute correlations can significantly degrade the model performance, especially on heterophilous graphs where connected nodes tend to have different labels~\cite{vcr}. As a result, these methods struggle to generate expressive neighborhood tokens that fully capture both topology and attribute correlations of nodes within the neighborhood, limiting their effectiveness for node classification.

\begin{figure}[t]
    \centering
        \subfigure[Photo]{
        \label{fig:1a}
        \begin{minipage}[t]{0.23\linewidth}
        \centering
        \includegraphics[scale=0.23]{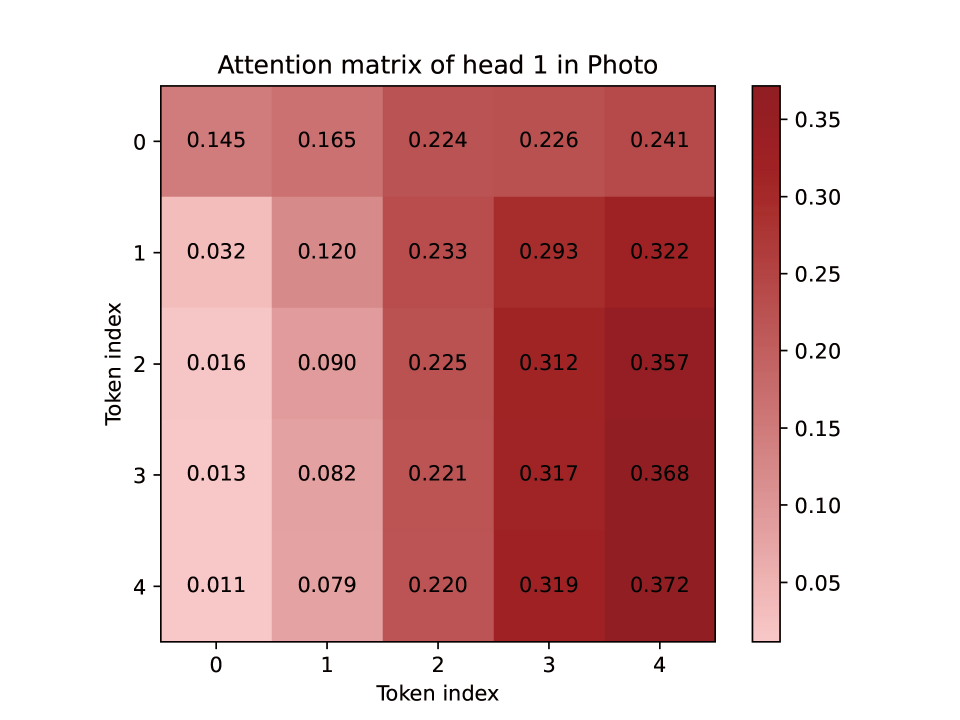}
        \end{minipage}
        
        \begin{minipage}[t]{0.23\linewidth}
        \centering
        \includegraphics[scale=0.23]{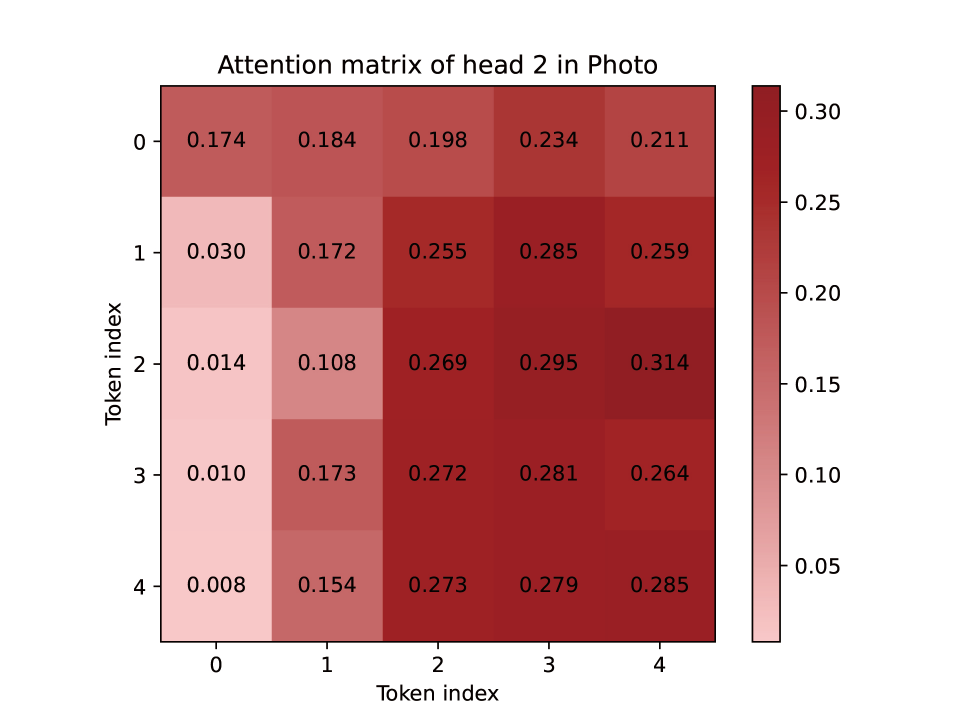}
        \end{minipage}
        }
        \subfigure[Reddit]{
        \label{fig:1b}
        \begin{minipage}[t]{0.23\linewidth}
        \centering
        \includegraphics[scale=0.23]{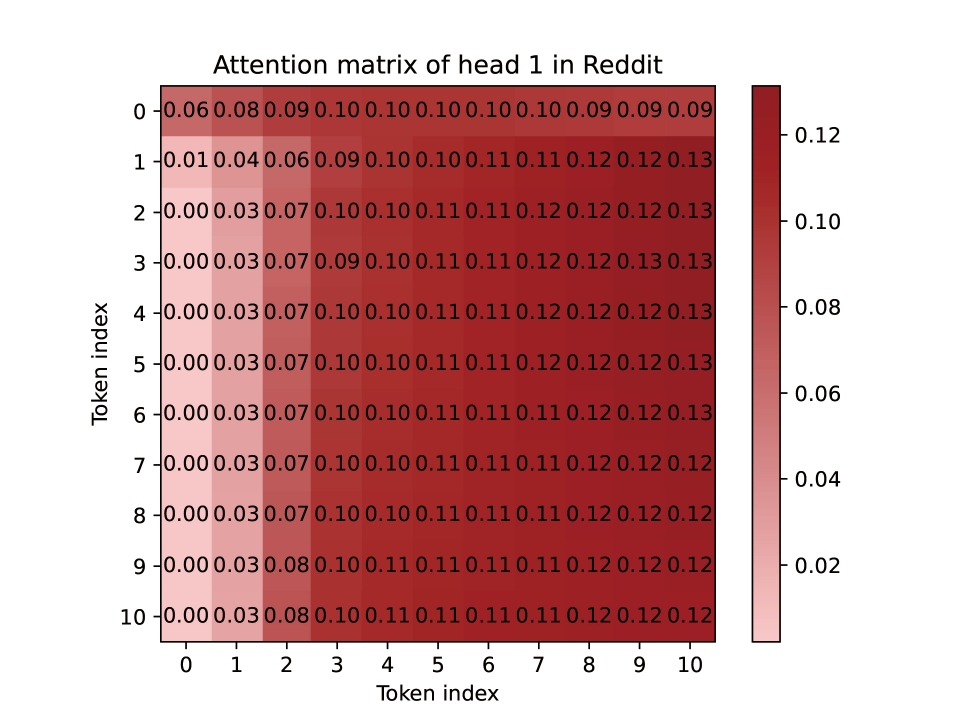}
        \end{minipage}
        
        \begin{minipage}[t]{0.23\linewidth}
        \centering
        \includegraphics[scale=0.23]{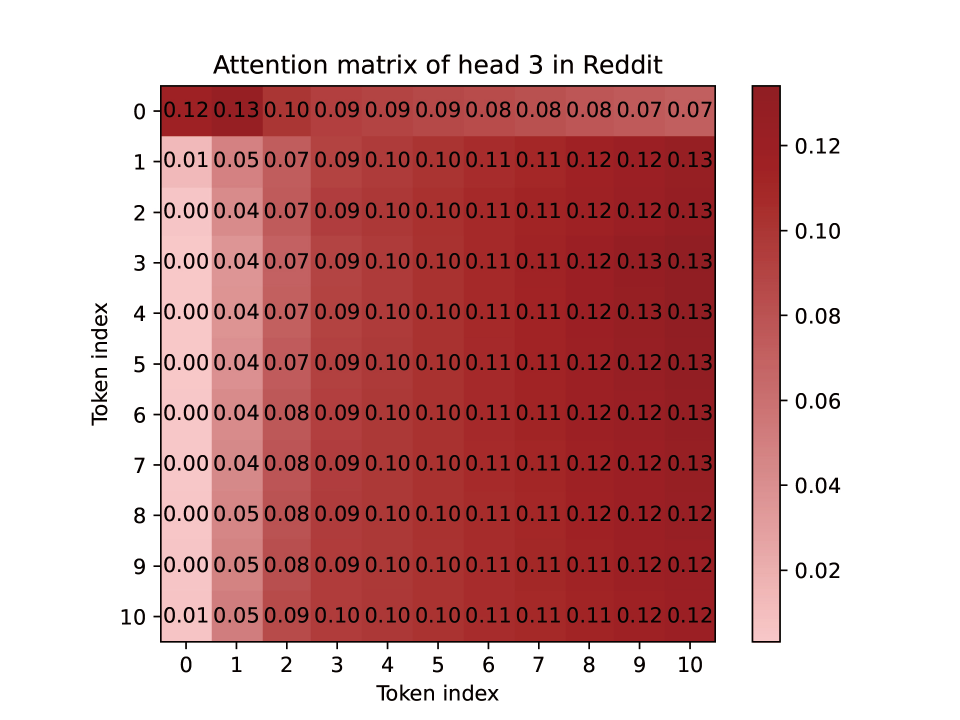}
        \end{minipage}
        }%
    \caption{The attention matrices on Photo and Reddit datasets. Deeper colors represent higher attention values.}
    \label{fig:vis_attention}
    \vspace{-1em}
\end{figure}

\textbf{(2) Suffering from the attention-diverting interference in self-attention.}
Through a visualization study of the attention matrix among neighborhood tokens (shown in \autoref{fig:vis_attention}, detailed discussions refer to Section \ref{DA}), we observe that the self-attention mechanism suffers from the attention-diverting interference issue when modeling neighborhood tokens, which describes that high-hop neighborhoods attract excessive attention, resulting in very low attention scores towards the target node. 
Consequently, the information interactions between the target node and its multi-hop neighborhoods are significantly weakened, undermining the goal of learning expressive node representations from different hop neighborhoods. 
As a result, the issue of attention-diverting interference severely hampers the potential of neighborhood-aware graph Transformers for graph representation learning.

To address these limitations, we propose \textbf{D}ual positional encoding-based \textbf{A}ttention \textbf{M}asking \textbf{G}raph \textbf{T}ransformer (\name), a novel graph Transformer model for node classification.
Specifically, to tackle limitation (1), \name introduces a novel dual positional encoding that incorporates a new positional encoding in the attribute feature space, based on attribute feature clustering.
By combining this attribute-aware positional encoding with the topology-aware positional encoding, the proposed dual positional encoding effectively captures both topology and attribute correlations of nodes within the neighborhood, resulting in more informative neighborhood tokens.
To address limitation (2), \name introduces a mask-aware self-attention mechanism based on a simple yet effective masking strategy, designed to strengthen the information flow between the target node and neighborhood tokens.
Our approach addresses the attention-diverting interference issue, thereby enhancing model performance for node classification. 
We conduct experiments on graphs of varying scales, ranging from thousands to millions of nodes, to validate the effectiveness of DAM-GT in node classification. The results show that DAM-GT significantly outperforms existing state-of-the-art methods. 

The main contributions are summarized as follows:

\begin{itemize}
\item We develop a new dual positional encoding that preserves both topology and attribute correlations of nodes into neighborhood tokens.

\item We observe the attention-diverting interference issue in existing self-attention mechanism when modeling neighborhood tokens, and further propose a novel mask-aware self-attention mechanism to effectively address this issue.
 
\item We propose \name, a novel graph Transformer for node classification, and demonstrate its superiority over competitive baselines through extensive empirical results. 

\end{itemize}

%% file: body/03preliminary.tex
Suppose an attributed graph $\mathcal{G}=\{V, E, \mathbf{X}, \mathbf{Y}\}$, where $V$ and $E$ are sets of nodes and edges in the graph $\mathcal{G}$.
$\mathbf{X}\in \mathbb{R}^{n\times d}$ is the attribute feature matrix.
$n$ is the number of nodes and $d$ is the dimension of feature vectors.
We also have the adjacency matrix $\mathbf{A}=\{0,1\}^{n\times n}$ and its normalized version $\hat{\mathbf{A}}=(\mathbf{D}+\mathbf{I})^{-1/2}(\mathbf{A}+\mathbf{I})(\mathbf{D}+\mathbf{I})^{-1/2}$, where $\mathbf{D}_{ii}=\sum_{j=1}^{n} \mathbf{A}_{ij}$ is 
the diagonal degree matrix.
$\mathbf{Y}\in \mathbb{R}^{n\times c}$ denotes the label matrix where each node is associated with a one-hot label vector to determine its label's information and $c$ is the number of classes.
Given a set of labeled nodes $V_L$, the goal of node classification is to predict the labels of rest nodes $V-V_L$.

%% file: body/04analysis.tex
Previous graph Transformers~\cite{nagformer,vcr} utilize the propagation-based strategy to construct neighborhood-aware tokens for each target node from the input graph.
Each token is generated by information aggregation of nodes within the specific neighborhood.
Due to the property of the graph structural data, there is an inclusion relation between obtained tokens, \ie high-hop neighborhoods contain nodes and information from low-hop neighborhoods.
Such an inclusion relation of neighborhood-aware tokens is significantly different from tokens in natural language processing and computer vision tasks where by no means exists above inclusion relation in a token sequence.

Intuitively, the specific inclusion relation between neighborhood-aware tokens could have a potential impact on the self-attention mechanism of Transformer since this module is used to capture the semantic relations of input tokens.
To investigate the influence of the inclusion relation on the calculation of attention scores, we adopt NAGphormer~\cite{nagformer} as the backbone and visualize the attention matrix in the well-trained model.
Due to the space limitation, we select Photo and Reddit for experiments to show the attention scores on small-scale and large-scale graphs, respectively. 
Considering the scales of graphs, we set the length of token sequences as 5 and 11 on Photo and Reddit, which means 4-hop neighborhood and 10-hop neighborhood are the max-hop neighborhoods in the token sequences (the target node itself is the 0-hop neighborhood), on Photo and Reddit respectively.
On each dataset, the four-head self-attention is adopted for model training.

The attention matrices are shown in \autoref{fig:vis_attention}. Deeper colors mean higher attention scores.
We can have the following observations:
(1) The target node (the first row in each attention matrix) generally has a similar level of attention towards itself and all neighborhoods.
(2) The neighborhood tokens (other rows in the attention matrix) have similar distributions of attention scores that tend to obtain large attention scores on high-hop neighborhood tokens and almost overlook the information of the target node (refer to the first column in each attention matrix).
Here we only show the attention matrix in two attention heads for each dataset due to the space limitation. Please refer to Appendix \ref{app:vis_atten_nag} for other results which exhibit the similar observations.

The above observations indicate that neighborhood tokens pay very little attention to the target node, and the corresponding attention distribution is diverted to high-hop neighborhood tokens.
We call this situation as the \textbf{attention-diverting interference issue}, which means that the representation updating of all neighborhood tokens heavily relies on high-hop neighborhoods which may contain much irrelevant even noisy information, almost ignoring the target node itself.
Rethinking the design of neighborhood-aware tokenized graph Transformer, the key idea is to learn node representations from multi-hop neighborhoods via the self-attention mechanism, which involves necessary information interactions between the target node and its multi-hop neighborhoods.
Unfortunately, the attention-diverting interference issue largely weakens the information interactions from the target node to neighborhood tokens, causing a serious impact on node representation learning.
Hence, this phenomenon motivates us to develop a new self-attention mechanism that ensures the effective information interactions between the target node and its multi-hop neighborhood tokens to address the attention-diverting interference issue in neighborhood-aware tokenized graph Transformers.

%% file: body/05method.tex
In this section, we elaborate on our proposed model, \name which encompasses two pivotal components: dual positional encoding and mask-aware self-attention mechanism. 
The overall framework of \name is shown in Figure \ref{fig:backbone}.

\subsection{Dual Positional Encoding}
Previous neighborhood tokens~\cite{nagformer} enhanced by topology-aware positional encoding are inefficient in capturing attribute-orient graph properties, such as heterophily~\cite{vcr}, which inevitably weakens the modeling ability of capturing diverse graph information.
To tackle this issue, we first introduce a novel attribute-aware positional encoding to effectively capture the relations of nodes in the attribute feature space. Specifically, we first apply the K-means algorithm to obtain $k$ clusters, $C_1,\dots, C_k$, ($k$ is the number of nodes' classes) according to the raw attribute features of nodes. The centroid representation $\mathbf{X}_{i}^{C}$ of a cluster $C_i$ is calculated as follows:
\begin{equation}
    \mathbf{X}_{i}^{C} = \frac{1}{|C_i|}\cdot \sum_{v_j\in C_i} \mathbf{X}_j.
    \label{eq:att_central}
\end{equation}
Then, the attribute-aware positional encoding $\mathbf{X}^{ap}_{j}$ of node $v_j$ belonging to cluster $C_i$ is calculated as follows:
\begin{equation}
    \mathbf{X}^{ap}_{j} = \delta_{v_j} \cdot \mathbf{X}_{i}^{C},
    \label{eq:ap_encoding}
\end{equation}
where $\delta_{v_j}$ denotes the correlation score between the target node and the centroid of the cluster. Naturally, a higher correlation score, a more similar representation.
In this paper, we adopt the cosine similarity to calculate $\delta_{v_j} = \mathrm{Cosine}(\mathbf{X}_j, \mathbf{X}_{i}^{C})$. 

Eq. (\ref{eq:ap_encoding}) ensures that nodes within the same cluster can obtain more similar positional encoding representations than those in different clusters.
Moreover, Eq. (\ref{eq:ap_encoding}) introduces the design of distance measurement $\delta$ to describe the relative positional relation between two nodes in the same cluster.
Therefore, the proposed attribute-aware positional encoding can completely preserve the both absolute and relative positional relations of nodes in the attribute feature space. 

Besides the proposed attribute-aware positional encoding, we also incorporate the topology-aware positional encoding to construct the final dual positional encoding:
\begin{equation}
    \mathbf{X}^{dup} = \mathbf{X}^{ap}||\mathbf{X}^{tp},
    \label{bi:ap_encoding}
\end{equation}
where $\mathbf{X}^{tp}\in \mathbb{R}^{n\times m}$ is the eigenvector matrix corresponding to the $m$ smallest non-trivial eigenvalues~\cite{nagformer} and $||$ is the concatenation operator.
$\mathbf{X}^{dup}\in \mathbb{R}^{n\times (d+m)}$ is the dual positional encoding matrix of nodes.

\subsection{Mask-aware Self-attention Mechanism}

As discussed in Section \ref{DA}, the original self-attention mechanism suffers from the attention-diverting interference issue, which hinders the information interactions between the target node and its neighborhood tokens, further weakening the model performance.
To tackle this limitation in the current self-attention mechanism, we develop a novel mask-aware self-attention mechanism, which leverages a simple yet effective masking strategy to guide the calculation of attention scores, ensuring the effective information learning from multi-hop neighborhood tokens.

Suppose $\mathbf{H}\in \mathbb{R}^{(S+1)\times d_{in}}$ is the input sequence of multi-hop neighborhoods where $S$ is the largest hop of neighborhood. The attention matrix of original self-attention mechanism is calculated as follows:
\begin{equation}
    \mathbf{M} = \frac{\mathbf{Q}\cdot \mathbf{K}^{\mathrm{T}}}{\sqrt{d_a}},
    \label{eq:ori-attention}
\end{equation}
where $\mathbf{Q}=\mathbf{H}\cdot \mathbf{W}_{Q}$ and $\mathbf{K}=\mathbf{H}\cdot \mathbf{W}_{K}$. $\mathbf{W}_{Q}\in \mathbb{R}^{d_{in}\times d_{a}}$ and $\mathbf{W}_{K}\in \mathbb{R}^{d_{in}\times d_{a}}$ are learnable parameter matrices and $d_{a}$ denotes the hidden dimension in the self-attention mechanism.
$\mathbf{M}\in \mathbb{R}^{(S+1)\times (S+1)}$ is the attention matrix that controls the interactions between multi-hop neighborhoods.

Based on the observations and discussions in Section \ref{DA}, we introduce a specific masking operation into the self-attention mechanism to address the attention-diverting interference issue.
Specifically, we first define a masking operator $\phi(\cdot)$ which will return the negative infinity when using the operator on any real number, denoted as:
\begin{equation}
\forall a\in \mathbb{R},\ \phi \left( a \right) =-\infty.
\end{equation}

Then, the masking operation is performed as follows:

\begin{equation}
\mathbf{\tilde{M}}_{i,j}=\left\{ \begin{array}{l}
	 \mathbf{M}_{i,j}, i*j=0 \ or \ i=j \\
	\phi \left( \mathbf{M}_{i,j} \right),others\\
\end{array} \right., 
\label{eq:mask-attention}
\end{equation}
where $i,j \in \{0,1,\dots, S\}$ denotes the index of row and column, respectively.
The final mask-aware attention matrix $\mathbf{M}^{\prime}$ is calculated as follows:
\begin{equation}
\mathbf{M}^{\prime} = \mathrm{softmax}( \mathbf{\tilde{M}} ).
\label{eq:mask-attention-2}
\end{equation}

Here, we keep the values in the first row, the first column, and diagonal of the original attention matrix $\mathbf{M}$ unchanged, and set values in other positions to the negative infinity. After the softmax operation, positions with negative infinity value become $0$. Therefore, non-zero values only exist in the first row, the first column, and diagonal of the mask-aware attention matrix $\mathbf{M}^{\prime}\in \mathbb{R}^{(S+1)\times (S+1)}$. 

We provide some insights for better understanding our designs:
The first row and the first column ensure the information interactions between the target node and its multi-hop neighborhoods.
And the diagonal reserves the information of tokens themselves during the update of representations.
These three components are crucial and necessary for learning informative node representations from their neighborhood tokens.
Therefore, we retain these components through the masking strategy to ensure the efficient and effective information flow between nodes and their neighborhood tokens, naturally avoiding the attention-diverting interference issue. 
We further conduct experiments to explore the contributions of these components on model performance, please refer to \autoref{app:mask_ablation}.

Finally, the output of the mask-aware self-attention mechanism is calculated as follows:
\begin{equation}
    \mathbf{H}^{\prime} = \mathbf{M}^{\prime} \cdot \mathbf{V},
    \label{eq:att-output}
\end{equation}
where $\mathbf{V}=\mathbf{H}\cdot \mathbf{W}_{V}$ and $\mathbf{W}_{V}\in \mathbb{R}^{d_{in}\times d_{a}}$ is the learnable parameter matrix.
The multi-head version called mask-aware multi-head self-attention mechanism (MMA) is calculated as follows:
\begin{equation}
\text{MMA}(\mathbf{H})=( \underset{h=1}{\overset{H}{||}}\mathbf{M}^{\prime h}\ \cdot \mathbf{V}^{h} ) \cdot\mathbf{W}^o,
\end{equation}
where $\mathbf{M}^{\prime h}$ is the mask-aware attention matrix of the $h$-th attention head.
$||$ is the concatenation operator and $H$ is the total number of attention heads. 
$\mathbf{W}^o \in \mathbb{R}^{(H\times d_{a})\times d_{out}}$ denotes the output linear transformation.

\subsection{\name for node classification}
Based on the proposed dual positional encoding and MMA, we further integrate these designs into the Transformer-based backbone to develop a novel method called \name for node classification.

Given the input graph $\mathcal{G}$, we first utilize Hop2Token \cite{nagformer} to generate neighborhood tokens.
Specifically, we combine the dual positional encoding with the raw attribute features to completely preserve the information of nodes within the same neighborhood.
The calculation of $s$-hop neighborhood token is as follows:
\begin{equation}
    \mathbf{N}^{(s)}=\hat{\mathbf{A}}^{s} \cdot \mathbf{X}^{\prime}, 
    \label{eq:hop-token}
\end{equation}
where $\hat{\mathbf{A}}^{s}$ denotes the $s$-order normalized adjacency matrix, $\mathbf{X}^{\prime}=\mathbf{X}||\mathbf{X}^{dup}$ denotes the enhanced feature matrix of nodes.
$\mathbf{N}^{(s)}\in \mathbb{R}^{n\times (2d+m)}$ denotes the $s$-hop neighborhood information of nodes and $\mathbf{N}^{(0)}=\mathbf{X}^{\prime}$.

For the target node $v$, we construct its neighborhood token sequence $\mathbf{T}^{v}\in \mathbb{R}^{(S+1)\times (2d+m)}$ as $\mathbf{T}^{v}=[\mathbf{N}^{(0)}_{v},\mathbf{N}^{(1)}_{v},\dots, \mathbf{N}^{(S)}_{v}]$.
Then, the projection layer is adopted to generate the model input of node $v$:
\begin{equation}
\mathbf{Z}^{(0)}_v =\mathbf{T}^{v} \cdot \mathbf{W}_p,
\end{equation}
where $\mathbf{W}_p\in \mathbb{R}^{(2d+m)\times d_m}$ indicates the trainable matrix of the linear projection and $\mathbf{Z}^{(0)}_v \in \mathbb{R}^{(S+1)\times d_m}$ denotes the input feature matrix of node $v$. 

Then, \name utilizes a series of modified Transformer layers by replacing MSA with MMA to learn node representations from the input token sequence:
\begin{equation}
\tilde{\mathbf{Z}}_{v}^{(l)}=\text{MMA}( \text{LN}( \mathbf{Z}_{v}^{( l-1 )} ) ) +\mathbf{Z}_{v}^{( l-1 )},
\end{equation}
\begin{equation}
\mathbf{Z}_{v}^{(l)}=\text{FFN}( \text{LN}( \tilde{\mathbf{Z}}_{v}^{( l )} ) ) +\tilde{\mathbf{Z}}_{v}^{(l)},
\end{equation}
where $l=1, \ldots, L$ indicates the $l$-th layer of \name.
$\text{LN}(\cdot)$ denotes the layer normalization.
$\text{FFN}(\cdot)$ denotes the standard feed-forward network in Transformer.
Through $L$ Transformer layers, the output ${\mathbf{Z}}_{v}^{(L)}\in \mathbb{R}^{(S+1)\times d_m}$ is the representations of all neighborhoods of node $v$.

To leverage the learned representations of multi-hop neighborhoods to predict the label of node $v$, \name utilizes the attention-based readout function~\cite{nagformer} to obtain the final representation of node $v$:
\begin{equation}
\omega _s=\frac{\text{exp}( ( \mathbf{Z}_{v,0}^{( L )}||\mathbf{Z}_{v,s}^{( L )} ) \mathbf{W}^\mathrm{T} )}{\sum\limits_{i=1}^S{\text{exp}( ( \mathbf{Z}_{v,0}^{( L )}||\mathbf{Z}_{v,i}^{( L )} ) \mathbf{W}^\mathrm{T} )}},
\end{equation}
\begin{equation}
\mathbf{Z}_{v}^{out}=\mathbf{Z}_{v,0}^{( L )}+\sum_{s=1}^S{\omega _s \mathbf{Z}_{v,s}^{( L )}},
\end{equation}
where $\mathbf{W}\in \mathbb{R}^{1\times 2d_m}$ is the trainable projection and $\omega _s$ is the aggregation weight of $s$-hop neighborhood token. $\mathbf{Z}_{v}^{out} \in \mathbb{R}^{d_m}$ is the final representation for node $v$. 
Finally, we adopt the Multilayer Perceptron and cross-entropy loss to obtain the predicted labels and train the model, which follows the same settings in previous studies~\cite{nagformer,nodeformer,sgformer}.
In addition, the detailed complexity analysis of \name is reported in Appendix \ref{app:complexity}.


%% file: body/06Exp.tex
\subsection{Datasets}
We conduct experiments on 12 widely used datasets of various scales and homophily levels, including nine small-scale datasets and three relatively large-scale datasets as well as homophilous and heterophilous graphs.
For small-scale datasets, we adopt six common benchmarks, \ie Pubmed, CoraFull, Computer, Photo, CS and Physics from the Deep Graph Library (DGL), as well as three publicly available datasets, \ie UAI2010, Flickr and BlogCatalog from~\cite{amgcn}. For large-scale datasets, we adopt three common datasets, \ie AMiner-CS, Reddit and Amazon2M from~\cite{grand+}.
The edge homophily ratio $H(\mathcal{G})\in [0,1]$ is adopted to measure the homophily level of graph $\mathcal{G}$.
Detailed descriptions of all datasets are summarized in Appendix~\ref{app:exp_set_data}.
We follow the split settings of large-scale datasets in~\cite{grand+}. For the other datasets, we apply 60\%/20\%/20\% train/val/test random splits.

\subsection{Baselines}
We select sixteen representative methods for the node classification task, including three categories of baselines, \ie MLP-based methods, GNNs, and graph Transformers. Specifically, for MLP-based baselines, we use SimMLP~\cite{simmlp} and ES-MLP~\cite{esmlp}. For GNNs, we adopt GCN~\cite{gcn}, GAT~\cite{gat}, APPNP~\cite{appnp}, GPRGNN~\cite{gprgnn}, LSGNN~\cite{LSGNN}, GraphSAINT~\cite{GraphSAINT}, PPRGo~\cite{PPRGo}, and GRAND+~\cite{grand+}.
And we select GraphGPS~\cite{gps}, NodeFormer~\cite{nodeformer}, SGFormer~\cite{sgformer}, ANS-GT~\cite{ansgt}, NAGphormer~\cite{nagformer} and VCR-Graphormer~\cite{vcr} as baseline graph Transformers.
The implementation details of all methods are summarized in Appendix~\ref{app:imple}.

\input{tab/smallrc}

\subsection{Comparisons on Small-scale Datasets}

We first validate the effectiveness of \name on nine small-scale graphs.
We run each model ten times with different random seeds and report the average accuracy and the corresponding standard deviation.
The results are shown in \autoref{tab:smallrc}.

We can observe that \name outperforms various GNN and graph Transformer baselines on all datasets consistently, demonstrating its effectiveness and superiority for node classification.
Moreover, \name beats representative and powerful neighborhood-aware tokenized graph Transformers NAGphormer and VCR-Graphormer on all datasets.
This is attributed to that the above methods suffer from the attention-diverting interference issue mentioned in Section \ref{DA}.
Besides, the lacking of attribute correlations in neighborhood tokens also weakens their performance on node classification.
Our proposed \name utilizes the novel dual positional encoding and mask-aware self-attention mechanism to successfully overcome the limitations in previous neighborhood-aware tokenized graph Transformers, leading to the promising model performance.

\input{tab/largerc}
\input{tab/mask_ablation}

\begin{figure}[h!]
    \centering    
    \includegraphics[width=0.95\linewidth]{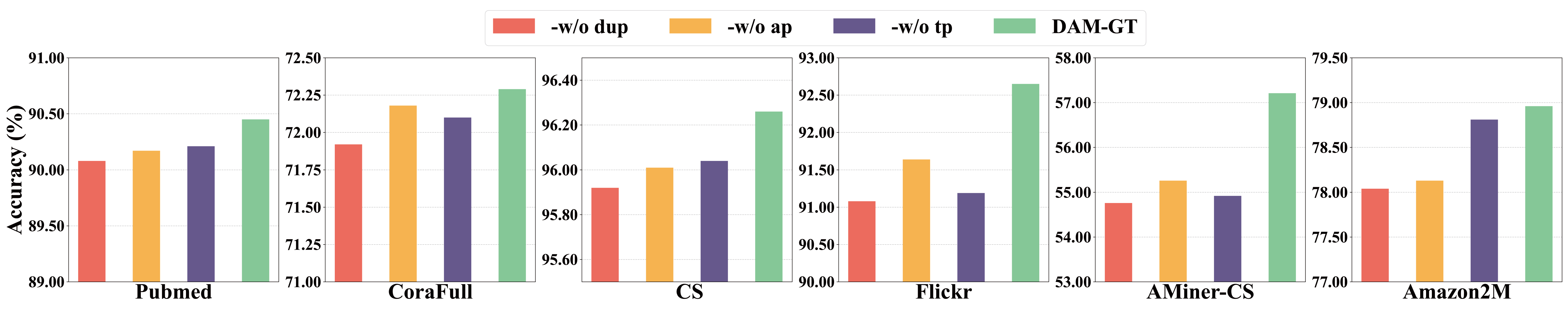}
    \caption{Comparison results of \name and its variants without positional encoding.}
    \label{fig:pe_ablation}
\end{figure}

\subsection{Comparisons on Large-scale Datasets}

To verify the superiority and scalability of \name on large-scale datasets, we next turn focus to comparisons on three large-scale datasets. 
As for the baselines, we only select scalable GNNs and scalable GTs because others suffer from the out-of-memory error due to their high computational cost. 
The comparison results are summarized in Table \ref{tab:largerc}. 
We can observe that our \name again outperforms all the competitors on all datasets consistently, demonstrating the desired scalability and excellent capacity for node classification in large graphs.
In addition, we conduct the efficiency study, following the settings in \cite{nagformer}.
The results are reported in Appendix~\ref{app:eff_and_sca} which indicate that \name exhibits competitive computing efficiency when handling large-scale graphs, compared to representative neighborhood-aware tokenized graph Transformers~\cite{nagformer,vcr}.

\subsection{Study on Mask-aware Self-attention Mechanism}
To validate the contribution of the mask-aware self-attention mechanism to the model performance, we develop a variant of \name by replacing it with the original self-attention mechanism.
Then we evaluate the performance of two models on all datasets.
The results are reported in Table \ref{tab:mask_ablation} where “-w/o mask” denotes the variant version.
We can observe that \name is superior to its variant on all graphs consistently, demonstrating the mask-aware attention mechanism plays a significant role in our \name. 
Moreover, to validate whether the proposed mask-aware self-attention mechanism addresses the attention-diverting interference issue, we further visualize the attention scores in \name. 
The results are shown in Appendix~\ref{app:vis_atten_dam}, which demonstrates that the proposed strategy can effectively overcome the above issue, resulting in improving model performance.

\subsection{Study on Dual Positional Encoding} 
To explore the influence of the positional encoding on model performance, we develop three variants of \name by removing dual positional encoding, attribute-aware positional encoding, and topology-aware positional encoding, named “-w/o dup”, “-w/o ap”, and “-w/o tp” respectively.  
We validate the performance of three variants on all datasets, and the results on six of them including small- and large-scale graphs are shown in \autoref{fig:pe_ablation} and results on other datasets are reported in Appendix~\ref{app:add_results_dpe}  due to the space limitation.

Based on the results, we can have the following observations:
1) \name consistently beats three variants on all datasets, showing that comprehensively considering the positional encoding in both topology and attribute feature spaces can effectively enhance the model performance for node classification;
2) The gains of applying single type of positional encoding vary on different datasets.
This is because different graphs exhibit varying properties, such as distinct attribute features or topology structures, resulting in different influences on model performance.
The results also suggest that capturing the positional relations between nodes in a single topology or attribute space is not enough for precisely preserving the complex graph properties and may undermine the generalization of the model.

\subsection{Study on Propagation Step}
The propagation step $S$, \ie the number of neighborhood tokens, is the key parameter of \name.
To further explore the influence of $S$ on model performance, we run \name with varying values of $S$ and observe the changes of model performance on all datasets.
Specifically, we vary $S$ in $\{2, \dots, 10\}$ and $\{4, \dots, 20\}$ for nine small-scale graphs and three large-scale graphs respectively due to the different scales of graphs. 
Similar to the previous section, we report results on six datasets in \autoref{fig:parameter} and other results are reported in Appendix~\ref{app:add_results_ps} which show similar observations.

\begin{figure}[t]
    \centering
    \includegraphics[width=0.95\linewidth]{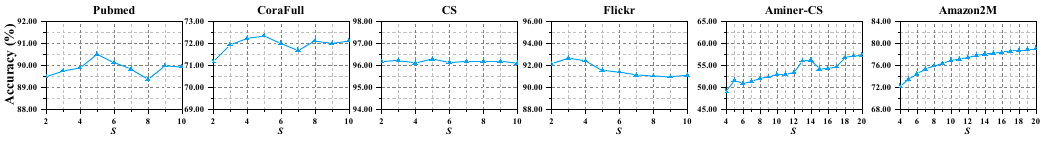}
    \vspace{-0.8em}
    \caption{Study on the propagation steps $S$.}
    \label{fig:parameter}
    \vspace{-1em}
\end{figure}

Generally speaking, the optimal value of $S$ is within five on small-scale graphs.
This is because that the whole graph may be connected within high-hop tokens on small-scale graphs.
In this situation, large-hop tokens cannot provide meaningful information for node representation learning.
Moreover, the change of model performance is less than 2\%, even within 0.3\% on CS.
This phenomenon indicates that \name is not sensitive to the over-smoothing issue.

And for large-scale graphs, the increasing $S$ can generally improve the model performance.
This situation reveals that large-scale graphs require a large value of $S$ to obtain more informative neighborhood tokens to learn high-quality node representations.
The results also indicate that the choice of $S$ is highly related to the scale of the input graph, and large-scale graphs prefer a large value of $S$ to achieve competitive performance.

%% file: tab/smallrc.tex
\begin{table*}[t]
\caption{Comparison results on small-scale datasets in terms of mean accuracy ± stdev (\%). The best results appear in bold.}
\centering
\label{tab:smallrc}
\scalebox{0.6}{
\begin{tabular}{lccccccccc}

\toprule
 Method & Pubmed     & CoraFull & Computer & Photo & CS & Physics & BlogCatalog & UAI2010 & Flickr \\ \midrule
SimMLP     &    87.85 ± 0.17 & 66.12 ± 0.21 & 88.92 ± 0.31 & 94.45 ± 0.25 & 95.43 ± 0.15 & 96.29 ± 0.04 &  93.32 ± 0.40 & 74.16 ± 0.72 & 88.08 ± 0.53 \\
ES-MLP    &   88.56 ± 0.23 & 63.78 ± 0.19 & 90.11 ± 0.34 & 93.95 ± 0.21 & 94.15 ± 0.11 & 96.27 ± 0.03 & 94.37 ± 0.55 & 76.45 ± 0.93 & 91.46 ± 0.82 \\  \midrule
GCN                        & 86.54 ± 0.12 &  61.76 ± 0.14   &  89.65 ± 0.52     &  92.70 ± 0.20   &  92.92 ± 0.12  &    96.18 ± 0.07 &  93.56 ± 0.43 & 74.68 ± 0.82 & 85.98 ± 0.64   \\
GAT  &     86.32 ± 0.16    &   64.47 ± 0.18 & 90.78 ± 0.13 & 93.87 ± 0.11 & 93.61 ± 0.14 & 96.17 ± 0.08  & 94.34 ± 0.64 & 75.17 ± 0.45 & 90.31 ± 0.37 \\
APPNP                      &     88.43 ± 0.15 & 65.16 ± 0.28 & 90.18 ± 0.17 & 94.32 ± 0.14 & 94.49 ± 0.07 & 96.54 ± 0.07 
& 94.21 ± 0.32 & 76.08 ± 0.66 & 89.97 ± 0.62   \\
GPRGNN                     &      89.34 ± 0.25 & 67.12 ± 0.31 & 89.32 ± 0.29 & 94.49 ± 0.14 & 95.13 ± 0.09 & 96.85 ± 0.08 & 95.02 ± 0.34 & 76.32 ± 0.59 & 90.48 ± 0.34  \\
LSGNN   &  89.73 ± 0.35 & 67.31 ± 0.41 & 90.42 ± 0.56 & 94.92 ± 0.30 & 95.19 ± 0.42 & 96.55 ± 0.07 & 95.23 ± 0.69 & 76.38 ± 1.04 & 91.52 ± 0.31 \\
GraphSAINT   &    88.96 ± 0.16 & 67.85 ± 0.21 & 90.22 ± 0.15 & 91.72 ± 0.13 & 94.41 ± 0.09 & 96.43 ± 0.05 & 94.34 ± 0.62 & 
74.47 ± 0.86 & 90.58 ± 0.62     \\
PPRGo           &    87.38 ± 0.11 & 63.54 ± 0.25 & 88.69 ± 0.21 & 93.61 ± 0.12 & 92.52 ± 0.15 & 95.51 ± 0.08 &94.05 ± 0.45 & 
76.51 ± 0.64 & 90.82 ± 0.96     \\ 
GRAND+                     &    88.64 ± 0.09 & 71.37 ± 0.11 & 88.74 ± 0.11 & 94.75 ± 0.12 & 93.92 ± 0.08 & 96.47 ± 0.04 & 
95.17 ± 0.42 & 77.25 ± 0.91 & 91.58 ± 0.58  \\ \midrule
GraphGPS                   &      88.94 ± 0.16 & 55.76 ± 0.23 & 90.98 ± 0.38 & 95.06 ± 0.13 & 93.93 ± 0.12  & 96.77 ± 0.06  & 95.63 ± 0.56
&77.45 ± 0.84
&88.64 ± 0.71        \\
NodeFormer   &     89.24 ± 0.14 & 61.82 ± 0.25 & 91.12 ± 0.19 & 95.27 ± 0.17 & 95.68 ± 0.08 & 97.19 ± 0.04  & 93.33 ± 0.85 & 73.87 ± 1.39 & 90.39 ± 0.49    \\
SGFormer   &     89.31 ± 0.17 & 65.32 ± 1.08 & 84.79 ± 0.70 & 92.43 ± 0.46 & 93.62 ± 0.05 & 96.71 ± 0.06  & 94.32 ± 0.21 & 75.17 ± 0.49 & 89.11 ± 1.04    \\
ANS-GT   &     89.36 ± 0.22 & 67.21 ± 0.34 & 90.47 ± 0.42 & 95.03 ± 0.24 & 95.17 ± 0.14 & 96.94 ± 0.06  & 94.18 ± 0.28 & 76.40 ± 0.48 & 89.95 ± 0.79   \\
NAGphormer       &     89.70 ± 0.19 & 71.51 ± 0.13 & 91.22 ± 0.14 & 95.49 ± 0.11 & 95.75 ± 0.09 & 97.34 ± 0.03  &  93.88 ± 0.64 & 78.05 ± 0.75 &  90.65 ± 0.61    \\
VCR-Graphormer    &   89.77 ± 0.15 & 71.67 ± 0.10 & 91.75 ± 0.15 & 95.53 ± 0.14 & 95.37 ± 0.04 & 97.34 ± 0.04    & 93.57 ± 0.42  & 77.51 ± 0.85  & 89.97 ± 0.78       \\ \midrule
\name         &    \textbf{90.45 ± 0.16} &  \textbf{72.29 ± 0.12}    &  \textbf{92.43 ± 0.18}   &  \textbf{96.66 ± 0.13}   &  \textbf{96.26 ± 0.08}   &  \textbf{97.40 ± 0.04}  &  \textbf{96.54 ± 0.30} & \textbf{79.67 ± 0.81} & \textbf{92.65 ± 0.41}      \\ \bottomrule
\end{tabular}
}
\end{table*}

%% file: tab/largerc.tex
\begin{table}[t]
    \caption{Comparison results on large-scale datasets in terms of mean accuracy ± stdev (\%). The best results appear in bold.}
    \label{tab:largerc}
\centering
\scalebox{0.7}{\begin{tabular}{lccc}
\toprule
Method & AMiner-CS  & Reddit & Amazon2M  \\ \midrule
GraphSAINT      &  51.86 ± 0.21  & 92.35 ± 0.08 & 75.21 ± 0.15  \\
PPRGo           &   49.07 ± 0.19 & 90.38 ± 0.11 & 66.12 ± 0.59  \\
GRAND+          &   54.67 ± 0.25 & 92.81 ± 0.03 & 75.49 ± 0.11  \\ \midrule
   
NodeFormer     &   44.75 ± 0.62 & 88.97 ± 0.32  & 71.56 ± 0.42 \\
SGFormer         &   48.52 ± 0.43 & 89.63 ± 0.26  & 74.22 ± 0.36 \\
ANS-GT         &   52.84 ± 0.74 & 93.23 ± 0.24  & 76.29 ± 0.32 \\
NAGphormer       &   56.21 ± 0.42  & 93.58 ± 0.05 & 77.43 ± 0.24  \\
VCR-Graphormer    &  53.59 ± 0.10 &  93.69 ± 0.08 & 76.09 ± 0.16   \\
\midrule
\name       &  \textbf{57.19 ± 0.35}     &    \textbf{93.92 ± 0.07}      &    \textbf{78.92 ± 0.33}  \\

\bottomrule
\end{tabular}
}
\end{table}

%% file: tab/mask_ablation.tex
\begin{table}[t]
    \caption{Comparison results of \name and the variant without the mask-aware self-attention mechanism on all datasets.}
    \label{tab:mask_ablation}
\centering
\scalebox{0.62}{
\begin{tabular}{ccccccccccccc} 

\toprule

 & Pubmed     & CoraFull & Computer & Photo & CS & Physics & BlogCatalog & UAI2010 & Flickr & AMiner-CS  & Reddit & Amazon2M   \\ \midrule
-w/o mask      &  89.92  & 71.82  & 91.80  & 96.15 & 95.91   & 97.23 & 95.89 & 78.54 & 91.44 & 56.41  & 93.78  & 78.51  \\
 \name     &  90.45 &  72.29  &  92.43 &  96.66   &  96.26  
 &  97.40  & 96.54  & 79.67  & 92.65 & 57.19 & 93.92  &  78.92  \\ \midrule
Gain      &  +0.53  & +0.47  & +0.63  & +0.51  & +0.35  & +0.17   & +0.65  & +1.13   & +1.21 &  +0.78 & +0.14 & +0.41   \\ 
\bottomrule
\end{tabular}
}
\end{table}

%% file: body/07conclusion.tex
In this paper, we identified two key limitations in existing neighborhood-aware tokenized graph Transformers: (1) the failure to capture attribute correlations among nodes within a neighborhood and (2) the attention-diverting interference issue in the self-attention mechanism. To address these challenges, we proposed \name, a scalable graph Transformer for node classification tasks.
Specifically, we introduced a dual positional encoding in \name that preserves both attribute and topology correlations of nodes in neighborhood tokens.
This enables the model to leverage more informative neighborhood tokens, enhancing its ability to learn expressive node representations. 
Additionally, to tackle the attention-diverting interference, we developed a mask-aware self-attention mechanism based on a simple yet effective masking strategy. 
By retaining necessary components of information interactions between the target node and its multi-hop neighborhoods in self-attention mechanism, the proposed mask-aware self-attention mechanism can effectively improve the information flow,
therefore facilitating the node representation learning from multi-hop neighborhood tokens. 
Extensive empirical results on a wide range of benchmark datasets showcase the superiority of \name in node classification when compared to representative GNNs and graph Transformers.

%% file: body/08APP.tex
\section{Related Work}
\label{app:rw}
\subsection{Graph Neural Network}
GNNs have been the effective approaches for the node classification task.
While, neighborhood aggregation is the most important module in a GNN-based method.
Prior GNNs~\cite{gcn,gat,graphsage} focus on aggregating the information of immediate neighbors.
In addition to immediate neighbors, follow-up methods aim to capture structural information beyond local neighborhood. 
A line of approaches~\cite{mixhop,jknet,gcnii} introduces the residual connection into GNNs.
Another line of methods~\cite{pt,pamt,ncgnn,appnp,sgc,gprgnn} adopts the decoupled design to achieve efficient information aggregation of multi-hop neighborhoods~\cite{rlp}. 

Besides aggregating neighborhood information based on the topology space, several studies~\cite{simpGNN,amgcn,geomgcn} explore the aggregation operation in the semantic feature space.
The key idea of aggregation operation in these methods is to aggregate the information of nodes with high similarity according to the input feature.

Previous GNN-based studies have shown that both topology and semantic features are important to learn informative node representations from the input graph.
This phenomenon motivates us to design a new positional encoding that not only captures the structural information, but also preserves the semantic information in the attribute feature space for effective node representation learning.

\subsection{Graph Transformer}
We divide existing graph Transformers for node classification in two groups according to the model input~\cite{ntformer}, \ie graph-based approaches and token-based approaches.

The former requires the graph as the input of Transformer. Approaches in this category develop various strategies, such as positional encoding~\cite{gt,gps} and GNN-style modules~\cite{gps,nodeformer,sgformer,gapformer,cob,dual}, to encode the graph structure information since Transformer regards the input graph as fully-connected, naturally ignoring the graph structural information. 

Though effective, a recent study~\cite{cob} has theoretically proved that directly applying the self-attention on the graph could cause the over-globalization issue, inevitably limiting the model performance.
Unlike previous methods, token-based approaches~\cite{gophormer,ansgt,nagformer,vcr} first transforms the input graph into short independent token sequences. Then, the generated token sequences are fed into Transformer to learn node representations.
This strategy naturally overcomes the over-globalization issue. Beside, the training cost is greatly reduced by applying mini-batch training strategy, effectively improving scalability to handle graphs of arbitrary scales which is important for node classification task.

Some methods select nodes with high similarity as tokens according to different sampling strategies, such as Top-$k$ sampling~\cite{ansgt} and random walk-based sampling~\cite{gophormer,ansgt}.
In addition to node-aware tokens, previous methods~\cite{nagformer,vcr} utilize the propagation strategies, such as random walk~\cite{nagformer} and personalized PageRank~\cite{vcr} to transform the multi-hop neighborhoods as token sequences.
Since neighborhood-aware tokens can comprehensively preserve the local topology information~\cite{vcr}, neighborhood token-based methods~\cite{nagformer,vcr} show promising performance in node classification.

Nevertheless, we observe that large hop neighborhood tokens can interfere with the distribution of attention scores due to the characteristics of neighborhood tokens.
This issue weakens the interactions between the target nodes and multi-hop neighborhoods, hindering from effectively learning node representations from neighborhood tokens.
Hence, we propose a new attention mechanism that adopts the masking strategy to ensure the effective information aggregation between the target node and its neighborhood tokens.

\begin{figure}[t]
    \centering
        \subfigure{
        \begin{minipage}[t]{0.23\linewidth}
        \centering
        \includegraphics[scale=0.238]{fig/Photo_red_1.pdf}
        \end{minipage}
        }%
        \subfigure{
        \begin{minipage}[t]{0.23\linewidth}
        \centering
        \includegraphics[scale=0.238]{fig/Photo_red_2.pdf}
        \end{minipage}
        }%
        \subfigure{
        \begin{minipage}[t]{0.23\linewidth}
        \centering
        \includegraphics[scale=0.238]{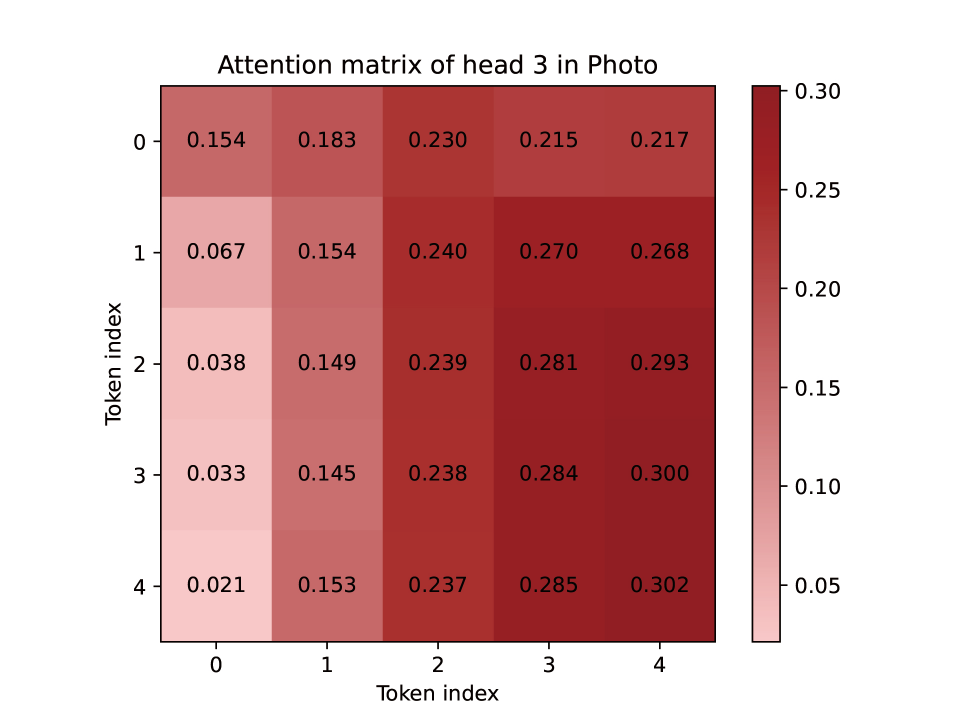}
        \end{minipage}
        }%
        \subfigure{
        \begin{minipage}[t]{0.23\linewidth}
        \centering
        \includegraphics[scale=0.238]{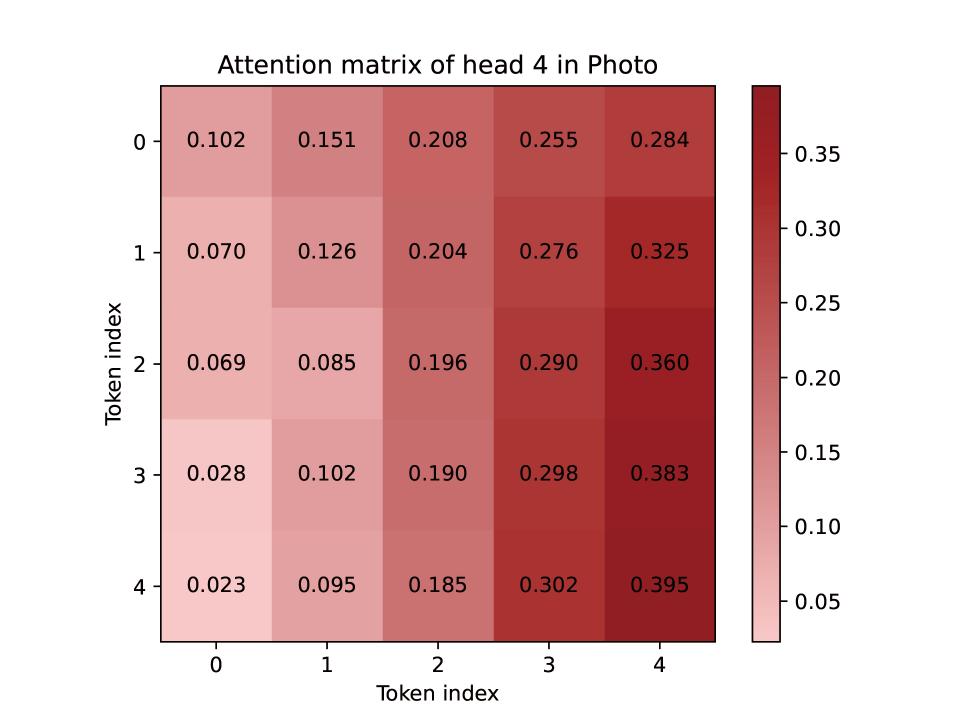}
        \end{minipage}
        }%
    \caption{The attention matrix of all four heads on Photo dataset in the backbone.}
    \label{fig:app_att1}
\end{figure}

\begin{figure}[t]
    \centering
        \subfigure{
        \begin{minipage}[t]{0.23\linewidth}
        \centering
        \includegraphics[scale=0.238]{fig/Reddit_red_1.pdf}
        \end{minipage}
        }%
        \subfigure{
        \begin{minipage}[t]{0.23\linewidth}
        \centering
        \includegraphics[scale=0.238]{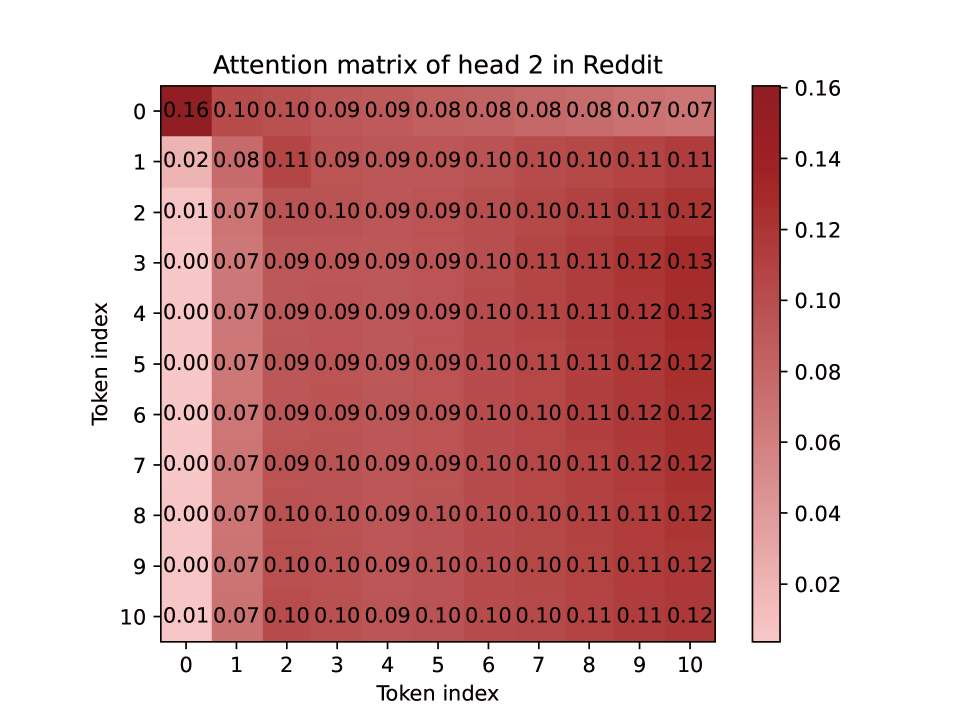}
        \end{minipage}
        }%
        \subfigure{
        \begin{minipage}[t]{0.23\linewidth}
        \centering
        \includegraphics[scale=0.238]{fig/Reddit_red_3.pdf}
        \end{minipage}
        }%
        \subfigure{
        \begin{minipage}[t]{0.23\linewidth}
        \centering
        \includegraphics[scale=0.238]{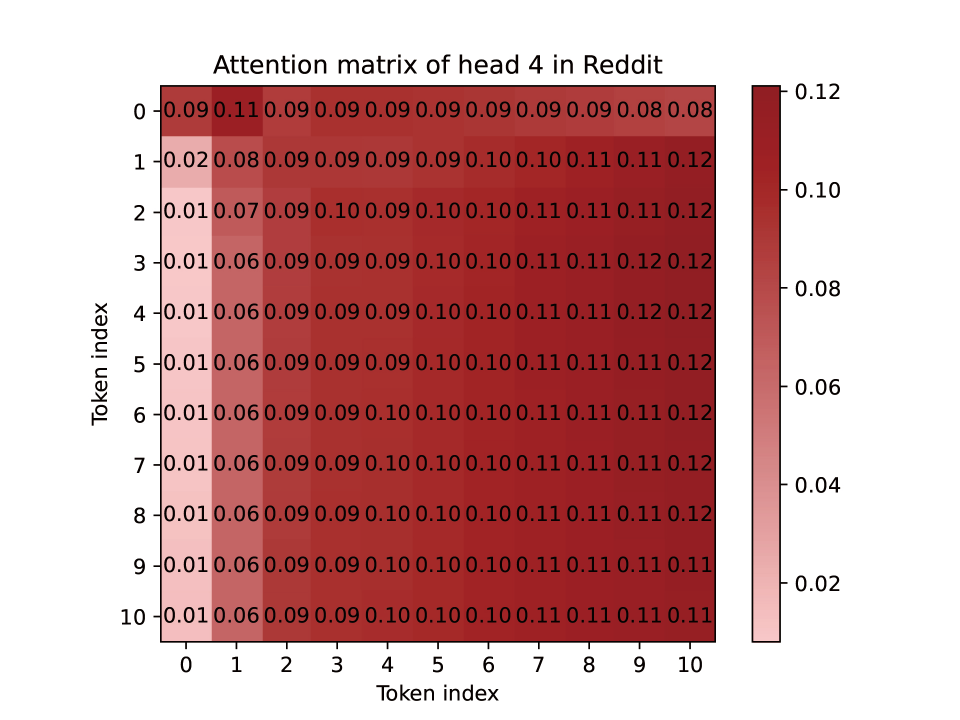}
        \end{minipage}
        }%
    \caption{The attention matrix of all four heads on Reddit dataset in the backbone.}
    \label{fig:app_att2}
\end{figure}

\begin{figure}[h!]
    \centering
        \subfigure{
        \begin{minipage}[h]{0.23\linewidth}
        \centering
        \includegraphics[scale=0.238]{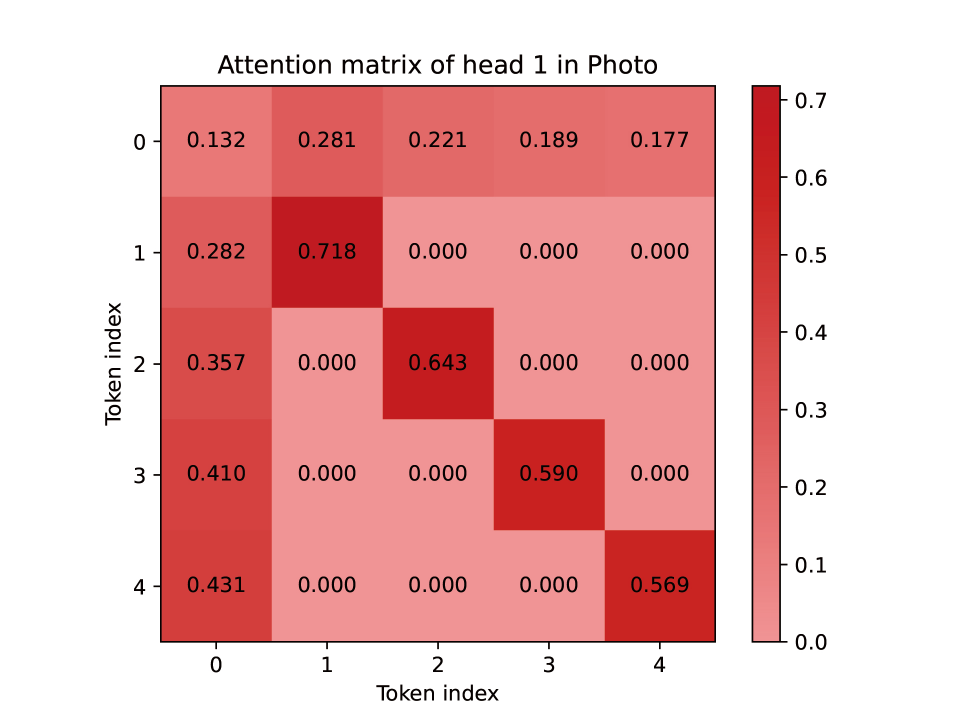}
        \end{minipage}
        }%
        \subfigure{
        \begin{minipage}[h]{0.23\linewidth}
        \centering
        \includegraphics[scale=0.238]{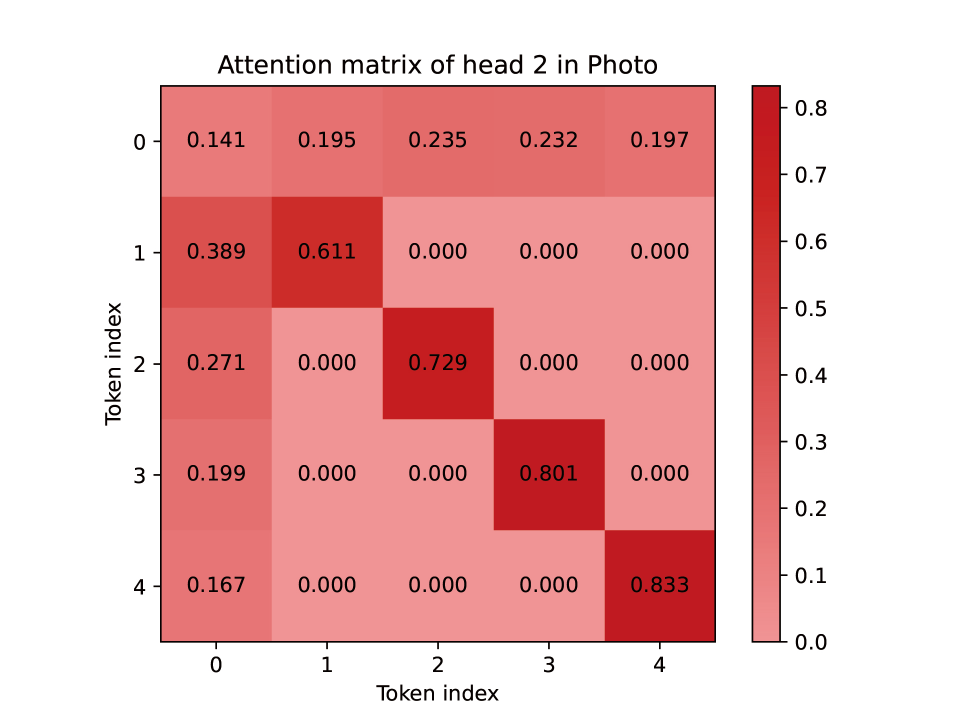}
        \end{minipage}
        }%
        \subfigure{
        \begin{minipage}[h]{0.23\linewidth}
        \centering
        \includegraphics[scale=0.238]{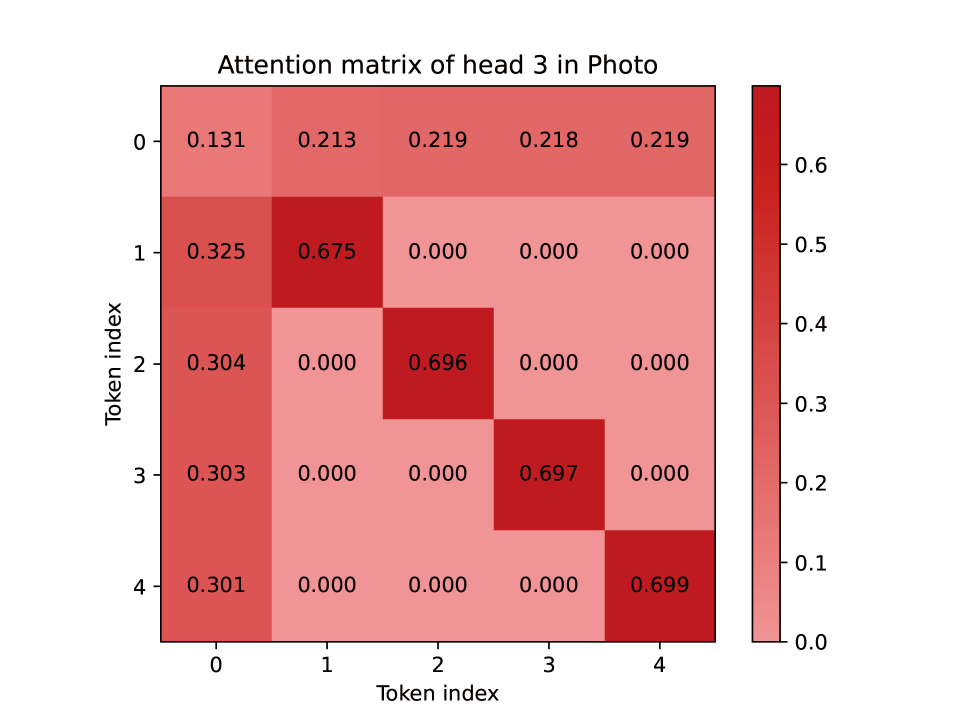}
        \end{minipage}
        }%
        \subfigure{
        \begin{minipage}[h]{0.23\linewidth}
        \centering
        \includegraphics[scale=0.238]{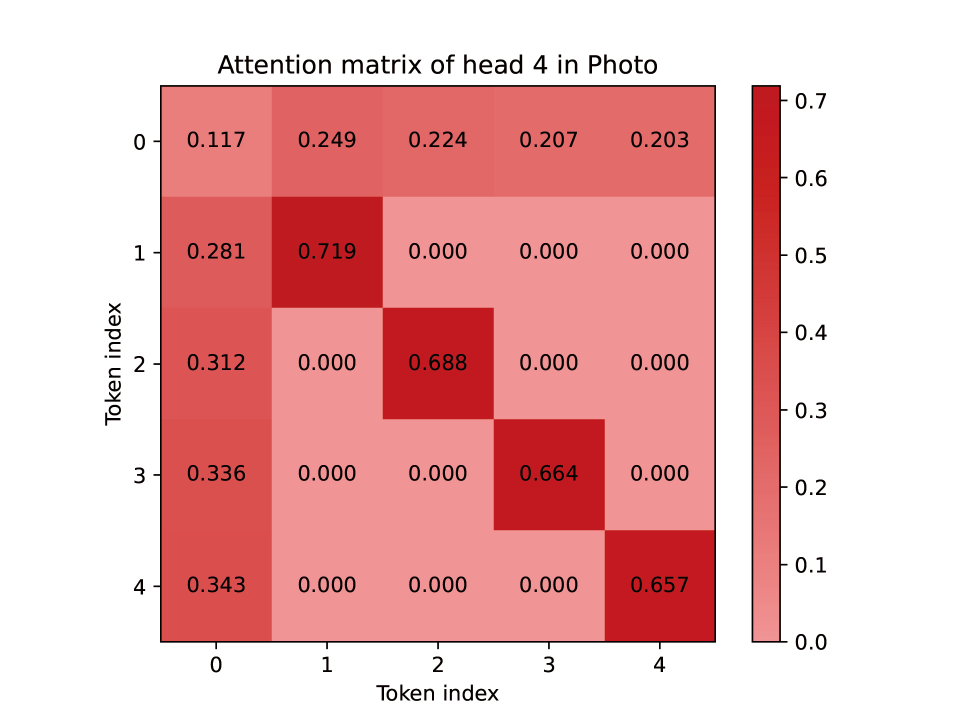}
        \end{minipage}
        }%
    \caption{The attention matrix of all four heads on Photo dataset in \name.}
    \label{fig:app_att3}
\end{figure}

\begin{figure}[h!]
    \centering
        \subfigure{
        \begin{minipage}[h]{0.23\linewidth}
        \centering
        \includegraphics[scale=0.238]{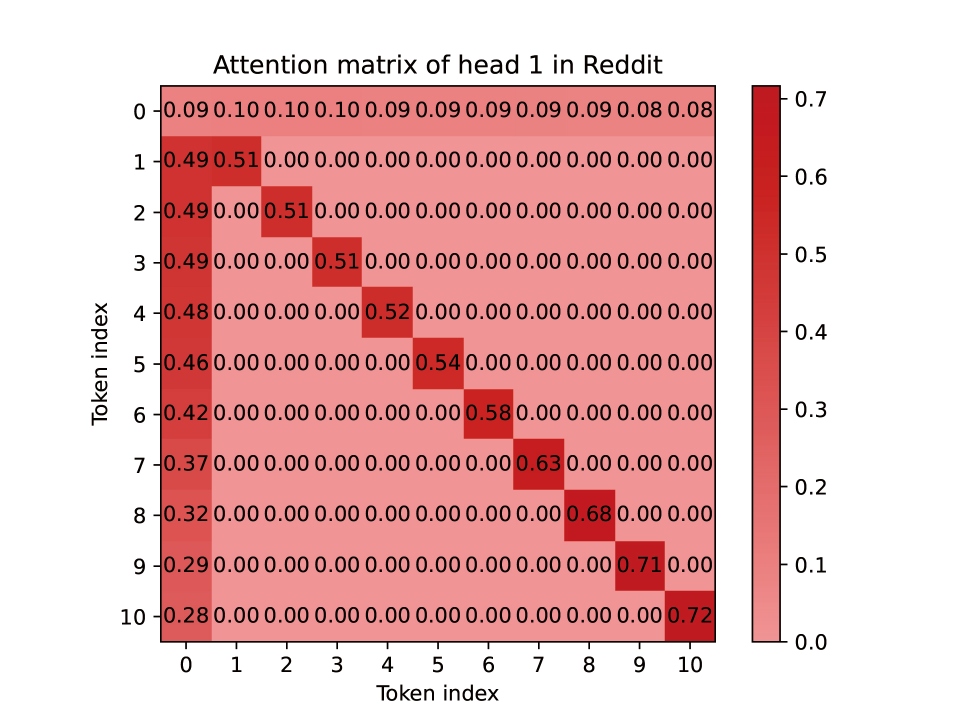}
        \end{minipage}
        }%
        \subfigure{
        \begin{minipage}[h]{0.23\linewidth}
        \centering
        \includegraphics[scale=0.238]{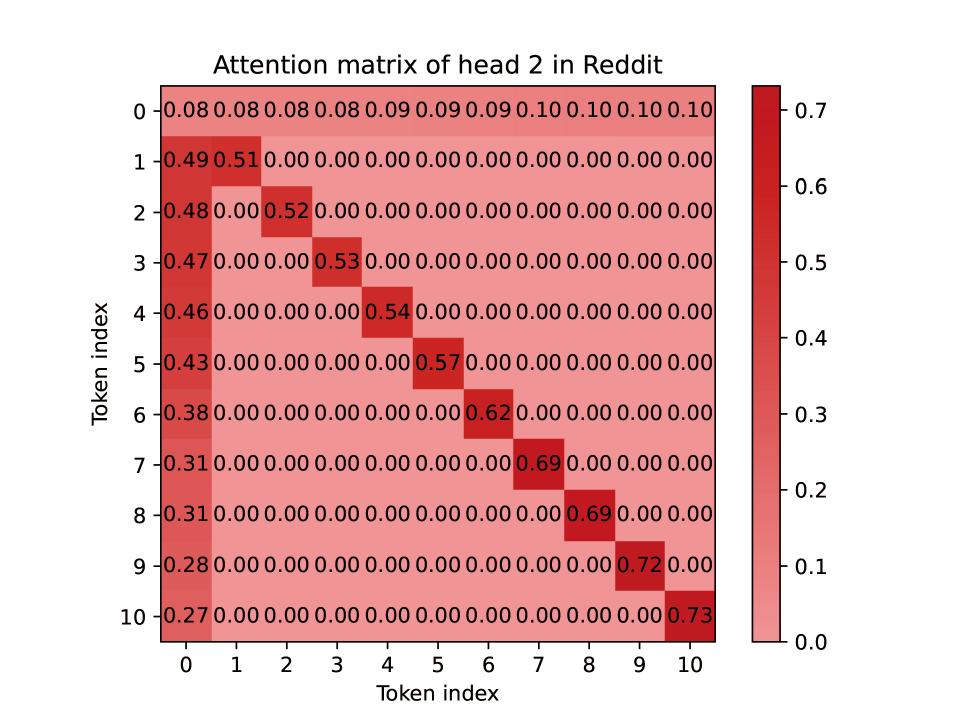}
        \end{minipage}
        }%
        \subfigure{
        \begin{minipage}[h]{0.23\linewidth}
        \centering
        \includegraphics[scale=0.238]{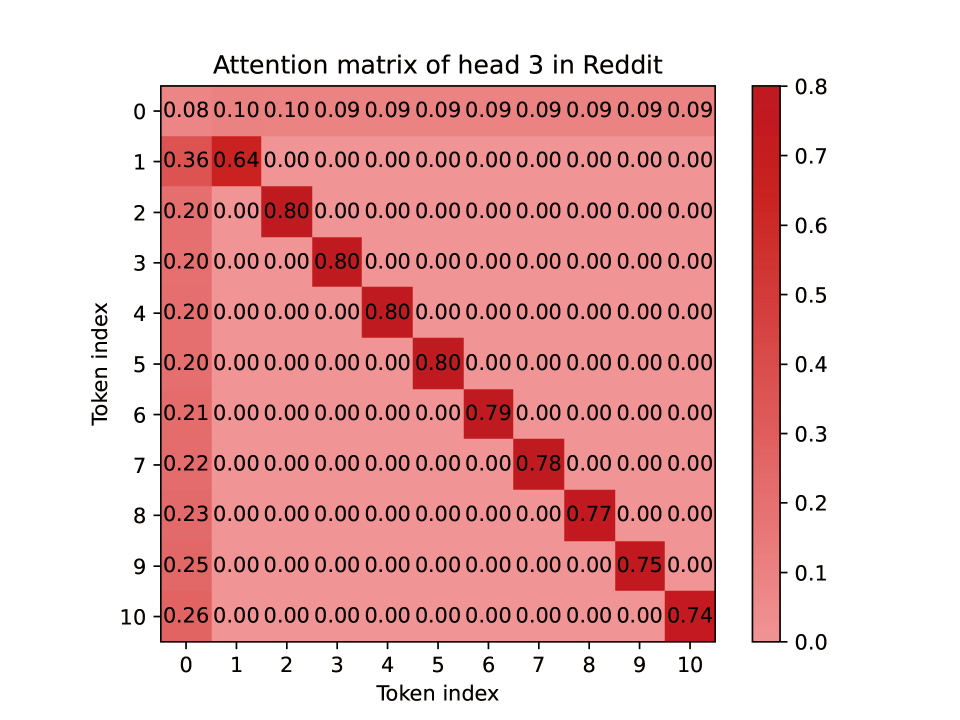}
        \end{minipage}
        }%
        \subfigure{
        \begin{minipage}[h]{0.23\linewidth}
        \centering
        \includegraphics[scale=0.238]{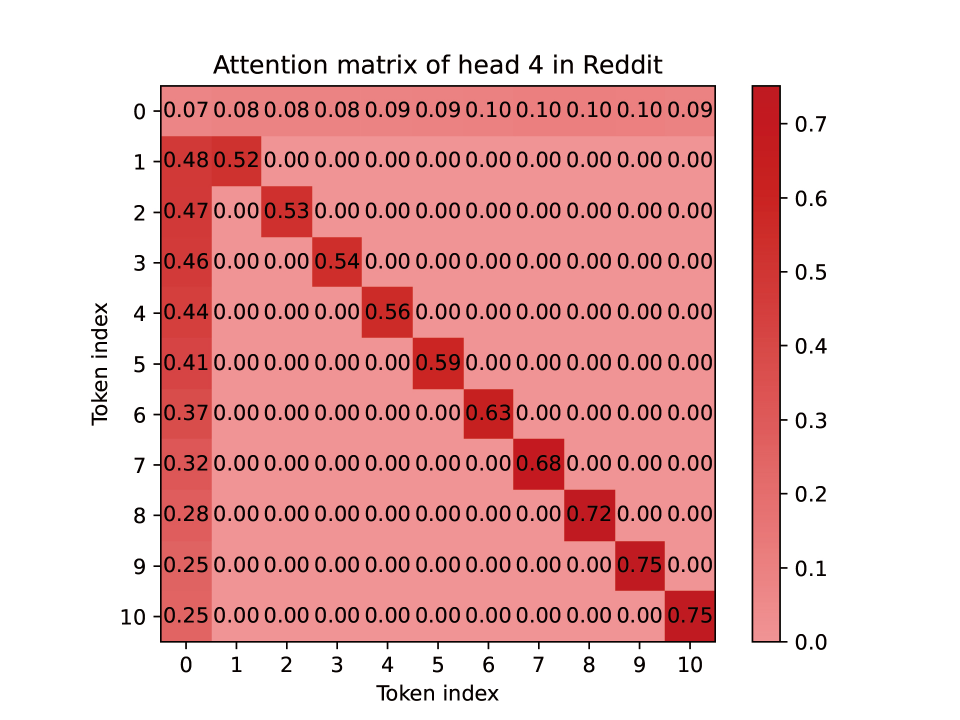}
        \end{minipage}
        }%
    \caption{The attention matrix of all four heads on Reddit dataset in \name.}
    \label{fig:app_att4}
\end{figure}

\section{Visualization of Attention Scores} \subsection{Attention Scores in the Backbone}\label{app:vis_atten_nag}
Here, we provide the attention score visualization of all attention heads on Photo and Reddit in the backbone.
The results are shown in \autoref{fig:app_att1} and \autoref{fig:app_att2}.
We can clearly observe that high-hop neighborhood token tend to attract excessive attention scores in each attention head on all datasets.
The phenomenon reveals that the original self-attention mechanism could suffer from the attention-diverting interference issue when dealing with the neighborhood token sequences.

\subsection{Attention Scores in \name}
\label{app:vis_atten_dam}
To figure out whether the attention-diverting interference issue is addressed in our \name, we keep consistent with the settings in Section \ref{DA}, and check the attention score matrix in our \name.
The attention score visualization of all four attention heads on Photo and Reddit in the proposed \name is shown in \autoref{fig:app_att3} and \autoref{fig:app_att4}.

We can observe that the attention scores of neighborhood tokens towards the target node in \name are normal values rather than anomalously weak values even zero, indicating that the attention-diverting interference issue is well addressed. Besides, it demonstrates the effectiveness of the proposed mask-aware attention mechanism.

\begin{figure}[t]
    \centering    
    \includegraphics[width=0.9\linewidth]{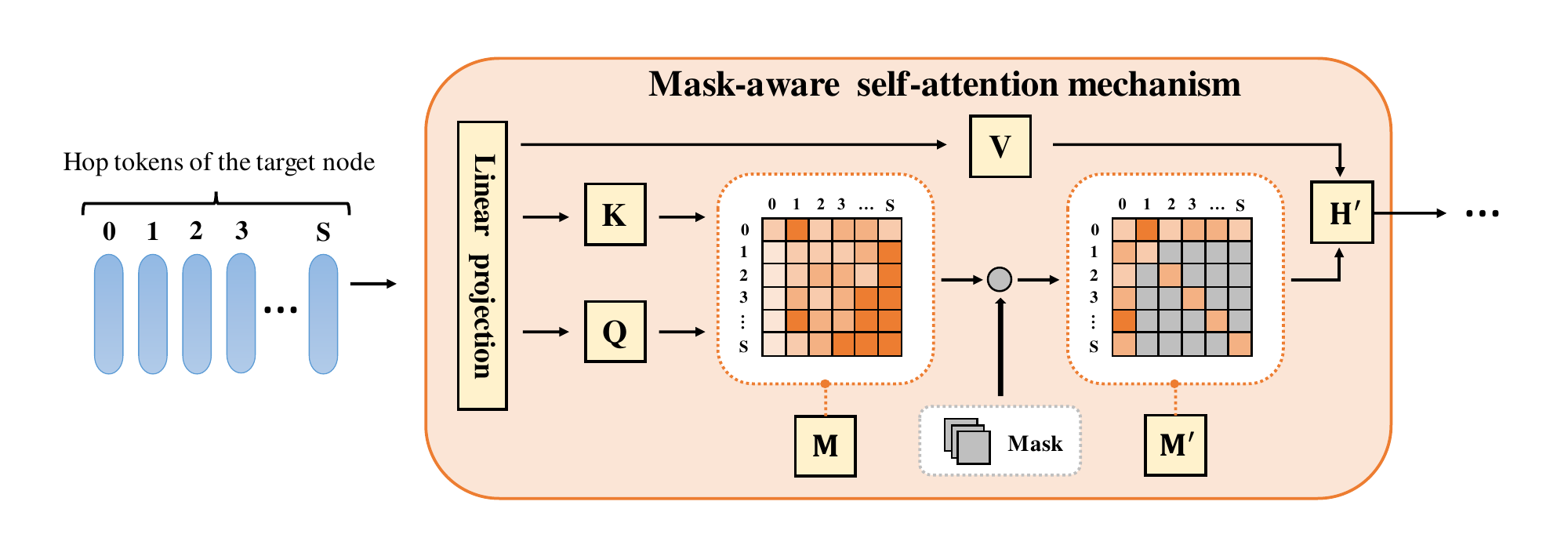}
    \caption{Illustration of the proposed mask-aware self-attention mechanism. To obtain the mask-aware self-attention matrix, we mask the positions except for the first row, the first column, and diagonal in the original attention matrix (with the mask marked in gray).
    }
    \label{fig:mask}
    \vspace{-1em}
\end{figure}

\begin{figure}[t]
    \centering
    \includegraphics[scale=0.8]{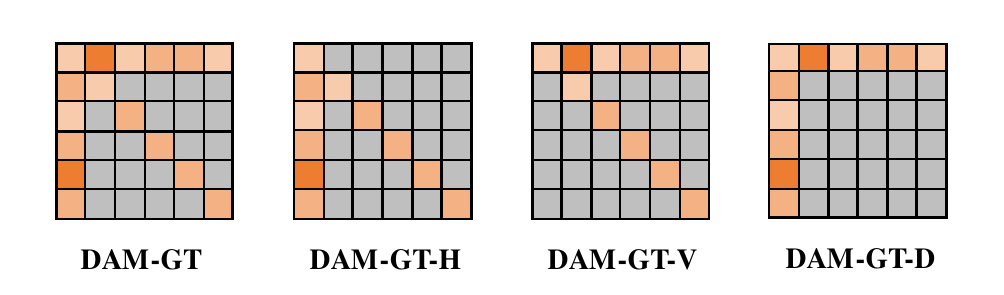}
    \caption{Masking strategies of \name and its three variants.}
    \label{fig:mask_strategy}
\end{figure}

\begin{figure}[t]
    \centering    
    \includegraphics[width=0.9\linewidth]{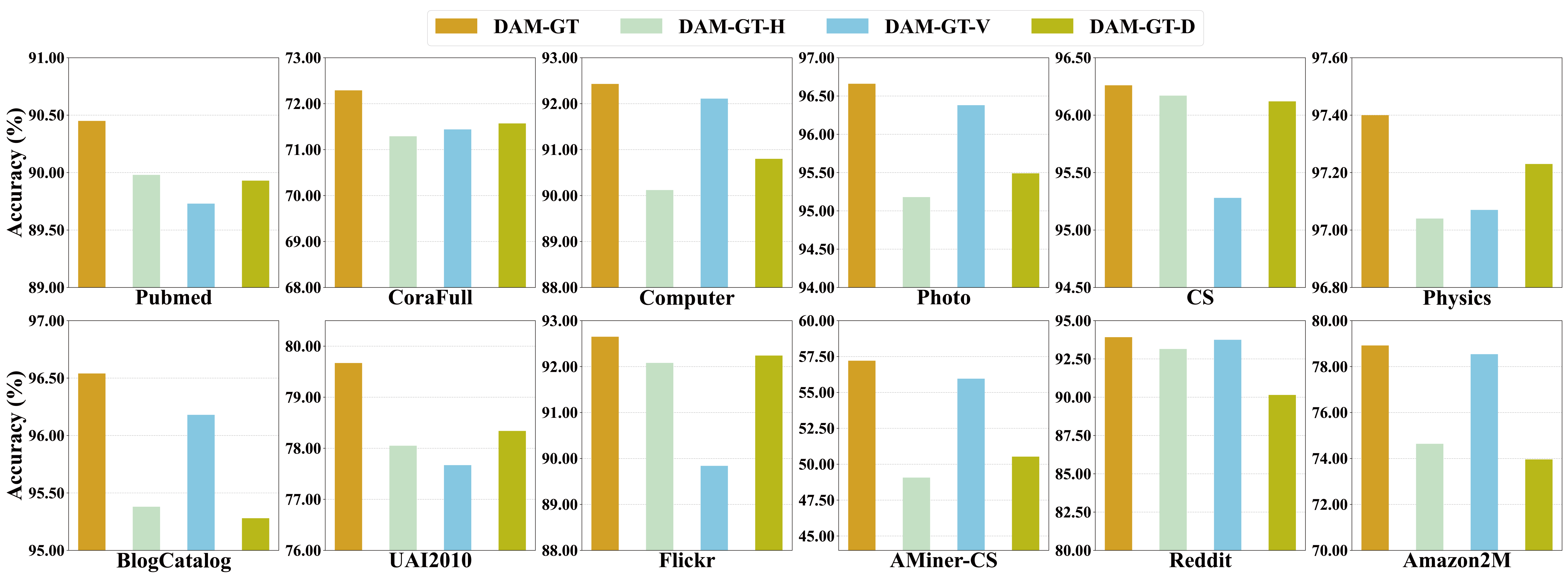}
    \caption{Comparison results of \name and its variants with different masking strategies. }
    \label{fig:mask_results}
\end{figure}

\section{Study on the Masking Strategy}
\label{app:mask_ablation}
Here, we provide detailed discussions and corresponding experiments to deeply understand our proposed masking strategy.

As described in Eq. (\ref{eq:mask-attention}), the rationale of the proposed masking strategy comes from two-fold:
(1) the target node $v$ can access itself and all multi-hop neighborhoods, learning information from different hops; 
(2) each neighborhood token can only access the target node $v$ and itself, effectively inducing neighborhood tokens to pay more attention to the target node.
In this way, the generated $\mathbf{M}^{\prime}$ can effectively capture the relations between the target node and its multi-hop neighborhoods, avoiding the attention-diverting interference issue. 

As shown in Figure \ref{fig:mask}, the proposed mask-aware self-attention mechanism actually only preserve the first row, the first column and the diagonal of the original attention matrix.
These components control the information interactions between the target node and its neighborhood tokens, further influencing the representation updating of them.

To deeply explore the influence of these key components on the model performance, we develop three variants of \name with different masking strategies, “DAM-GT-H”, “DAM-GT-V" and “DAM-GT-D". Figure \ref{fig:mask_strategy} visualizes the masking strategies in detail.
In “DAM-GT-H”, we retain the first column and the diagonal, which means we only preserve the information from the target node to neighborhood tokens.
“DAM-GT-V” retains the first row and the diagonal, which only preserves the information from neighborhood tokens to the target node.
The diagonal is retained in above two variants to preserve the information from tokens to themselves during the representation updating.
And “DAM-GT-D” retains the first row and the first column, which removes diagonal to verify the role of the residual structure.

We run three variants on all datasets and report the corresponding experimental results in Figure \ref{fig:mask_results}.
We can observe that \name consistently outperforms three variants on all datasets, demonstrating the superiority of our masking strategy.
Above comparisons also demonstrate that bidirectional information interactions between the target node and neighborhood tokens are both essential and ought to be preserved when designing masking strategy. Besides, the performance degradation due to the absence of diagonal indicates the importance of residual structure.
Moreover, these three variants exhibit distinct performance levels in different graphs, indicating that different graphs have varying demands for the three types of information.
The results also suggest our proposed masking strategy benefits from the three primary components, finally enhancing the model performance for node classification.

\section{Complexity Analysis of \name}
\label{app:complexity}

\subsection{Time Complexity}
Here, we discuss the time complexity from two stages: the pre-processing stage and the training stage. In pre-processing, the generation of topology and attribute position encoding requires $O(\left| E \right|^{3/2})$ and $O(Ikdn)$ respectively, and the calculation of neighborhood tokens requires $O(\left| E \right|)$. Here, $\left| E \right|$ is the number of edges, $I$ and $k$ are respectively the number of maximum iterations and the cluster number in k-means, $d$ is the dimension of node feature, and $n$
 is the number of nodes. In training stage, the time complexity is mainly from the attention module, which is $O(n(s+1)^2d_m)$, where $s$ is the number of hops and $d_m$ is the dimension of hidden vector in Transformer.

\subsection{Space Complexity}
The space complexity of our \name is $O(b(s+1)^2+b(s+1)d_m+d_m^2L)$ in the training stage, where $L$ denotes the number of Transformer layers and $b$ is the batchsize. We can find that the memory cost of training \name on the GPU device is restricted to the batchsize $b$. Therefore, if batchsize is suitable, \name can handle large-scale graphs even on limited GPU resources, showing good scalability.

\section{Experimental Setup}
\label{app:exp_set}

\input{tab/datasets}
\subsection{Datasets}\label{app:exp_set_data}
We adopt various graph benchmark datasets with different edge homophily levels as well as different graph scales.
The edge homophily level $H(\mathcal{G})$ is calculated as follows:
\begin{equation}
    {H}(\mathcal{G}) =  \frac{|\{e_{i,j} |(v_i,v_j)\in E, \mathbf{Y}_{i}=\mathbf{Y}_{j}\}|}{|E|}.
\end{equation}
High ${H}(\mathcal{G})$ means the graph exhibits strong homophily, implying that connected nodes tend to belong to the same label.
In contrast, low ${H}(\mathcal{G})$ means the graph exhibits strong heterophily, implying that connected nodes tend to belong to different labels.
The statistics of all datasets are reported in \autoref{tab:datasets}.

\subsection{Implementation Details}
\label{app:imple}

According to the recommended settings in the official implementations, we perform hyper-parameter tuning for each baseline. 
For the model configuration of \name, we try the number of Transformer layers in \{1, 2\}, the propagation steps in \{3, 4, ..., 20\}, and the hidden dimension in \{128, 256, 512, 768\}. Parameters are optimized with the AdamW~\cite{adamw} optimizer, using a learning rate of \{1e-2, 5e-3, 1e-3\} and a weight decay of \{1e-4, 5e-5, 1e-5\}. 
Besides, we set the batch size to be 2000. The number of heads $H$ in MMA is 8. The training process is early stopped within 50 epochs. All experiments are performed on a Linux machine with one NVIDIA RTX 2080Ti 11GB GPU.

\section{Efficiency Experiments}
\label{app:eff_and_sca}
In this section, we evaluate computational efficiency and scalability of \name on three large datasets, compared with the strong baselines NAGphormer and VCR-Graphormer. The training time (TT), inference time (IT), and GPU memory usage (MU) are reported in \autoref{tab:eff_and_sca}.

We can find that compared to NAGphormer, there is some additional time overhead due to the masking operation, which is affordable considering the highly improved accuracy. Moreover, compared to VCR-Graphormer, which is a follow-up method of NAGphormer, \name still exhibits high computational efficiency and scalability when handling large-scale graphs.

\input{tab/eff_and_sca}

\begin{figure*}[t]
    \centering    
    \includegraphics[width=0.9\linewidth]{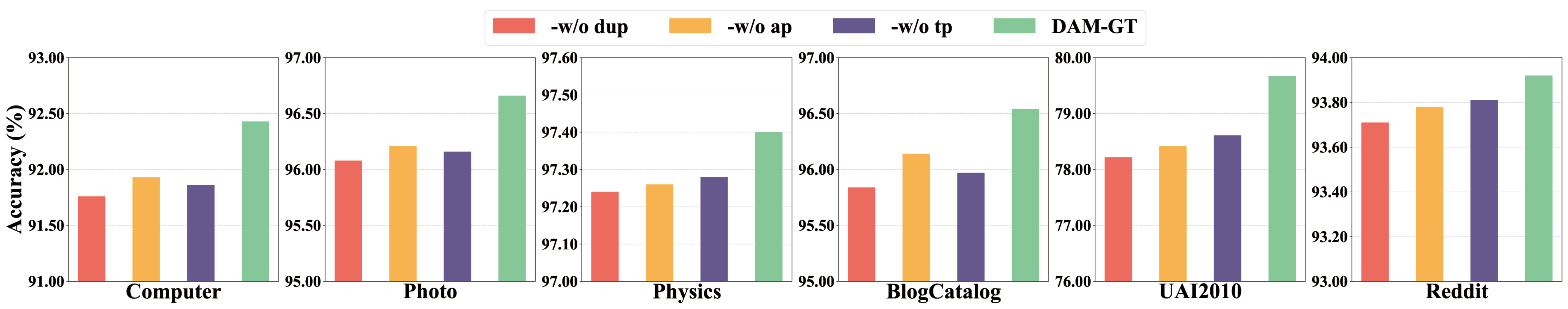}
    \caption{Comparison results of \name and its variants without positional encoding.}
    \label{app:pe_ablation}
\end{figure*}

\section{Additional Experimental Results}
\label{app:add_results}
\subsection{Study on Dual Positional Encoding}
\label{app:add_results_dpe}
The results of \name with different positional encoding strategies on other six datasets are provided in \autoref{app:pe_ablation}. 
We can find that the gains of single type of positional encoding vary on different datasets, which exhibits similar results with \autoref{fig:pe_ablation}.
In addition, \name with the proposed dual positional encoding consistently outperforms other variants on all datasets, showing the effectiveness and necessity of encoding the positional information of nodes in different feature spaces due to the complex graph information.

\subsection{Study on Propagation Step}
\label{app:add_results_ps}
\begin{figure}[t]
    \centering
    \includegraphics[width=0.95\linewidth]{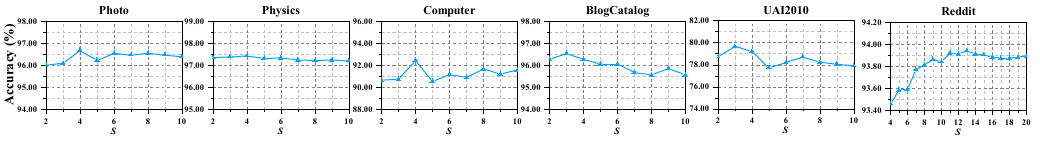}
 
    \caption{Performances of \name with various propagation steps $S$ on another six datasets.}
    \label{fig:parameter_appendix}
\end{figure}

Results of \name with different propagation steps are shown in \autoref{fig:parameter_appendix}.
We can clearly observe that small-scale graphs require a small value of $S$, while large-scale graphs prefer a large $S$.
This observation is consistent with that in the experiment section.
In summary, the choice of $S$ is strong related to the graph scale.
Based on the empirical results, the optimal value of $S$ is within five on graphs with thousands of nodes.
And on large-scale graphs with more than one hundred thousand nodes, the value of $S$ should be over ten to guarantee the model performance.

\section{Limitation of \name}\label{app:limitation}
Although the proposed mask-aware self-attention mechanism can effectively address the attention-diverting interference issue, it roughly blocks the interactions between other neighborhood tokens except the node itself and the target neighborhood token, which may cause the information loss of multi-hop neighborhoods.
One potential solution to address this limitation is to introduce a probability-based masking strategy which may preserve the information interactions between neighborhood tokens.

%% file: tab/datasets.tex
\begin{table}[t]
    \caption{Statistics of all datasets.}
    \label{tab:datasets}
    \centering
    \scalebox{1.0}{\begin{tabular}{l r r r r r}
    \toprule
    Dataset & \#Nodes $\uparrow$ & \#Edges & \#Features & \#Classes & $H(\mathcal{G})$  \\
    \midrule
    UAI2010  &  3,067 & 28,311 & 4,973 & 19 & 0.36 \\
     BlogCatalog  &   5,196 & 171,743 & 8,189 & 6 & 0.40 \\
    Flickr  &  7,575 & 239,738 & 12,047 & 9 & 0.24 \\
    Photo & 7,650 & 119,043 & 745 & 8 &  0.83 \\    
         Computer &  13,752 & 245,778 & 767 & 10 &  0.78 \\
     CS &   18,333 & 81,894 & 6,805 & 15 & 0.81 \\
   
    PubMed & 19,717 & 44,338 & 500 & 3  & 0.80 \\
    CoraFull & 19,793 & 65,311 & 8,710 & 70 &  0.57  \\
     Physics & 34,493 & 247,962 & 8,415 & 15 & 0.93 \\ 
         Reddit &  232,965 & 11,606,919 & 602 & 41 & 0.78 \\
    Aminer-CS &  593,486 & 6,217,004 & 100 & 18 & 0.63 \\

    Amazon2M  &  2,449,029 & 61,859,140 & 100 & 47 & 0.81 \\
    \bottomrule
    \end{tabular}
    }

\end{table}

%% file: tab/eff_and_sca.tex
\begin{table}[t]
    \caption{The comparison results on large-scale graphs in terms of training time (TT), inference time (IT), and GPU memory usage (MU).}
    \label{tab:eff_and_sca}
    \centering
    \scalebox{0.8}{
\begin{tabular}{cccccccccc}
\toprule
 & \multicolumn{3}{c}{Aminer-CS}   & \multicolumn{3}{c}{Reddit}   & \multicolumn{3}{c}{Amazon2M}      \\
  & TT (s) & IT (s) & MU (MB) & TT (s) & IT (s) & MU (MB) & TT (s) & IT (s) & MU (MB) \\ \midrule
NAGphormer     & 9.12              & 34.93              & 2504        & 19.55             & 10.26              & 1864        & 9.55              & 59.97              & 2514        \\
VCR-Graphormer & 11.45             & 46.83              & 3896        & 53.18             & 28.42              & 9506        & 13.47             & 127.58             & 20652       \\ \midrule
DAM-GT         & 10.18             & 43.52              & 2512        & 27.33             & 17.78              & 1932        & 12.19             & 88.36        & 2552    \\ \bottomrule  
\end{tabular}
}
\end{table}

%% file: main.bbl

\begin{thebibliography}{43}


\ifx \showCODEN    \undefined \def \showCODEN     #1{\unskip}     \fi
\ifx \showDOI      \undefined \def \showDOI       #1{#1}\fi
\ifx \showISBNx    \undefined \def \showISBNx     #1{\unskip}     \fi
\ifx \showISBNxiii \undefined \def \showISBNxiii  #1{\unskip}     \fi
\ifx \showISSN     \undefined \def \showISSN      #1{\unskip}     \fi
\ifx \showLCCN     \undefined \def \showLCCN      #1{\unskip}     \fi
\ifx \shownote     \undefined \def \shownote      #1{#1}          \fi
\ifx \showarticletitle \undefined \def \showarticletitle #1{#1}   \fi
\ifx \showURL      \undefined \def \showURL       {\relax}        \fi
\providecommand\bibfield[2]{#2}
\providecommand\bibinfo[2]{#2}
\providecommand\natexlab[1]{#1}
\providecommand\showeprint[2][]{arXiv:#2}

\bibitem[\protect\citeauthoryear{Abu{-}El{-}Haija, Perozzi, Kapoor, Alipourfard, Lerman, Harutyunyan, Steeg, and Galstyan}{Abu{-}El{-}Haija et~al\mbox{.}}{2019}]%
        {mixhop}
\bibfield{author}{\bibinfo{person}{Sami Abu{-}El{-}Haija}, \bibinfo{person}{Bryan Perozzi}, \bibinfo{person}{Amol Kapoor}, \bibinfo{person}{Nazanin Alipourfard}, \bibinfo{person}{Kristina Lerman}, \bibinfo{person}{Hrayr Harutyunyan}, \bibinfo{person}{Greg~Ver Steeg}, {and} \bibinfo{person}{Aram Galstyan}.} \bibinfo{year}{2019}\natexlab{}.
\newblock \showarticletitle{MixHop: Higher-Order Graph Convolutional Architectures via Sparsified Neighborhood Mixing}. In \bibinfo{booktitle}{\emph{Proceedings of the 36th International Conference on Machine Learning, {ICML}}}, Vol.~\bibinfo{volume}{97}. \bibinfo{pages}{21--29}.
\newblock


\bibitem[\protect\citeauthoryear{Alon and Yahav}{Alon and Yahav}{2021}]%
        {oversq}
\bibfield{author}{\bibinfo{person}{Uri Alon} {and} \bibinfo{person}{Eran Yahav}.} \bibinfo{year}{2021}\natexlab{}.
\newblock \showarticletitle{{On the Bottleneck of Graph Neural Networks and its Practical Implications}}. In \bibinfo{booktitle}{\emph{Proceedings of the International Conference on Learning Representations, {ICLR}}}.
\newblock


\bibitem[\protect\citeauthoryear{Bojchevski, Klicpera, Perozzi, Kapoor, Blais, R{\'o}zemberczki, Lukasik, and G{\"u}nnemann}{Bojchevski et~al\mbox{.}}{2020}]%
        {PPRGo}
\bibfield{author}{\bibinfo{person}{Aleksandar Bojchevski}, \bibinfo{person}{Johannes Klicpera}, \bibinfo{person}{Bryan Perozzi}, \bibinfo{person}{Amol Kapoor}, \bibinfo{person}{Martin Blais}, \bibinfo{person}{Benedek R{\'o}zemberczki}, \bibinfo{person}{Michal Lukasik}, {and} \bibinfo{person}{Stephan G{\"u}nnemann}.} \bibinfo{year}{2020}\natexlab{}.
\newblock \showarticletitle{{Scaling Graph Neural Networks with Approximate Pagerank}}. In \bibinfo{booktitle}{\emph{Proceedings of the ACM SIGKDD International Conference on Knowledge Discovery \& Data Mining}}. \bibinfo{pages}{2464--2473}.
\newblock


\bibitem[\protect\citeauthoryear{Chen, Lin, Li, Li, Zhou, and Sun}{Chen et~al\mbox{.}}{2020a}]%
        {oversmoothing}
\bibfield{author}{\bibinfo{person}{Deli Chen}, \bibinfo{person}{Yankai Lin}, \bibinfo{person}{Wei Li}, \bibinfo{person}{Peng Li}, \bibinfo{person}{Jie Zhou}, {and} \bibinfo{person}{Xu Sun}.} \bibinfo{year}{2020}\natexlab{a}.
\newblock \showarticletitle{{Measuring and Relieving the Over-smoothing Problem for Graph Neural Networks From the Topological View}}. In \bibinfo{booktitle}{\emph{Proceedings of the AAAI Conference on Artificial Intelligence, {AAAI}}}. \bibinfo{pages}{3438--3445}.
\newblock


\bibitem[\protect\citeauthoryear{Chen, Gao, Li, and He}{Chen et~al\mbox{.}}{2023a}]%
        {nagformer}
\bibfield{author}{\bibinfo{person}{Jinsong Chen}, \bibinfo{person}{Kaiyuan Gao}, \bibinfo{person}{Gaichao Li}, {and} \bibinfo{person}{Kun He}.} \bibinfo{year}{2023}\natexlab{a}.
\newblock \showarticletitle{{NAGphormer: A Tokenized Graph Transformer for Node Classification in Large Graphs}}. In \bibinfo{booktitle}{\emph{Proceedings of the International Conference on Learning Representations, {ICLR}}}.
\newblock


\bibitem[\protect\citeauthoryear{Chen, Jiang, and He}{Chen et~al\mbox{.}}{2024a}]%
        {ntformer}
\bibfield{author}{\bibinfo{person}{Jinsong Chen}, \bibinfo{person}{Siyu Jiang}, {and} \bibinfo{person}{Kun He}.} \bibinfo{year}{2024}\natexlab{a}.
\newblock \showarticletitle{NTFormer: {A} Composite Node Tokenized Graph Transformer for Node Classification}.
\newblock \bibinfo{journal}{\emph{CoRR}}  \bibinfo{volume}{abs/2406.19249} (\bibinfo{year}{2024}).
\newblock


\bibitem[\protect\citeauthoryear{Chen, Li, and He}{Chen et~al\mbox{.}}{2024b}]%
        {ncgnn}
\bibfield{author}{\bibinfo{person}{Jinsong Chen}, \bibinfo{person}{Boyu Li}, {and} \bibinfo{person}{Kun He}.} \bibinfo{year}{2024}\natexlab{b}.
\newblock \showarticletitle{Neighborhood convolutional graph neural network}.
\newblock \bibinfo{journal}{\emph{Knowledge-Based Systems}}  \bibinfo{volume}{295} (\bibinfo{year}{2024}), \bibinfo{pages}{111861}.
\newblock


\bibitem[\protect\citeauthoryear{Chen, Li, He, and He}{Chen et~al\mbox{.}}{2024c}]%
        {pamt}
\bibfield{author}{\bibinfo{person}{Jinsong Chen}, \bibinfo{person}{Boyu Li}, \bibinfo{person}{Qiuting He}, {and} \bibinfo{person}{Kun He}.} \bibinfo{year}{2024}\natexlab{c}.
\newblock \showarticletitle{{PAMT:} {A} Novel Propagation-Based Approach via Adaptive Similarity Mask for Node Classification}.
\newblock \bibinfo{journal}{\emph{{IEEE} Trans. Comput. Soc. Syst.}} \bibinfo{volume}{11}, \bibinfo{number}{5} (\bibinfo{year}{2024}), \bibinfo{pages}{5973--5983}.
\newblock


\bibitem[\protect\citeauthoryear{Chen, Wei, Huang, Ding, and Li}{Chen et~al\mbox{.}}{2020b}]%
        {gcnii}
\bibfield{author}{\bibinfo{person}{Ming Chen}, \bibinfo{person}{Zhewei Wei}, \bibinfo{person}{Zengfeng Huang}, \bibinfo{person}{Bolin Ding}, {and} \bibinfo{person}{Yaliang Li}.} \bibinfo{year}{2020}\natexlab{b}.
\newblock \showarticletitle{{Simple and Deep Graph Convolutional Networks}}. In \bibinfo{booktitle}{\emph{Proceedings of the International Conference on Machine Learning, {ICML}}}. \bibinfo{pages}{1725--1735}.
\newblock


\bibitem[\protect\citeauthoryear{Chen, Luo, Tang, Yang, Qiu, Wang, and Cao}{Chen et~al\mbox{.}}{2023b}]%
        {LSGNN}
\bibfield{author}{\bibinfo{person}{Yuhan Chen}, \bibinfo{person}{Yihong Luo}, \bibinfo{person}{Jing Tang}, \bibinfo{person}{Liang Yang}, \bibinfo{person}{Siya Qiu}, \bibinfo{person}{Chuan Wang}, {and} \bibinfo{person}{Xiaochun Cao}.} \bibinfo{year}{2023}\natexlab{b}.
\newblock \showarticletitle{{LSGNN:} Towards General Graph Neural Network in Node Classification by Local Similarity}. In \bibinfo{booktitle}{\emph{Proceedings of the Thirty-Second International Joint Conference on Artificial Intelligence, {IJCAI}}}. \bibinfo{pages}{3550--3558}.
\newblock


\bibitem[\protect\citeauthoryear{Chien, Peng, Li, and Milenkovic}{Chien et~al\mbox{.}}{2021}]%
        {gprgnn}
\bibfield{author}{\bibinfo{person}{Eli Chien}, \bibinfo{person}{Jianhao Peng}, \bibinfo{person}{Pan Li}, {and} \bibinfo{person}{Olgica Milenkovic}.} \bibinfo{year}{2021}\natexlab{}.
\newblock \showarticletitle{{Adaptive Universal Generalized PageRank Graph Neural Network}}. In \bibinfo{booktitle}{\emph{Proceedings of the International Conference on Learning Representations, {ICLR}}}.
\newblock


\bibitem[\protect\citeauthoryear{Dong, Chen, Feng, He, Bi, Ding, and Cui}{Dong et~al\mbox{.}}{2021}]%
        {pt}
\bibfield{author}{\bibinfo{person}{Hande Dong}, \bibinfo{person}{Jiawei Chen}, \bibinfo{person}{Fuli Feng}, \bibinfo{person}{Xiangnan He}, \bibinfo{person}{Shuxian Bi}, \bibinfo{person}{Zhaolin Ding}, {and} \bibinfo{person}{Peng Cui}.} \bibinfo{year}{2021}\natexlab{}.
\newblock \showarticletitle{{On the Equivalence of Decoupled Graph Convolution Network and Label Propagation}}. In \bibinfo{booktitle}{\emph{Proceedings of the Web Conference, {WWW}}}. \bibinfo{pages}{3651--3662}.
\newblock


\bibitem[\protect\citeauthoryear{Dwivedi and Bresson}{Dwivedi and Bresson}{2020}]%
        {gt}
\bibfield{author}{\bibinfo{person}{Vijay~Prakash Dwivedi} {and} \bibinfo{person}{Xavier Bresson}.} \bibinfo{year}{2020}\natexlab{}.
\newblock \showarticletitle{{A Generalization of Transformer Networks to Graphs}}.
\newblock \bibinfo{journal}{\emph{arXiv preprint arXiv:2012.09699}} (\bibinfo{year}{2020}).
\newblock


\bibitem[\protect\citeauthoryear{Feng, Dong, Huang, Yin, Cheng, Kharlamov, and Tang}{Feng et~al\mbox{.}}{2022}]%
        {grand+}
\bibfield{author}{\bibinfo{person}{Wenzheng Feng}, \bibinfo{person}{Yuxiao Dong}, \bibinfo{person}{Tinglin Huang}, \bibinfo{person}{Ziqi Yin}, \bibinfo{person}{Xu Cheng}, \bibinfo{person}{Evgeny Kharlamov}, {and} \bibinfo{person}{Jie Tang}.} \bibinfo{year}{2022}\natexlab{}.
\newblock \showarticletitle{{GRAND+: Scalable Graph Random Neural Networks}}. In \bibinfo{booktitle}{\emph{Proceedings of the Web Conference, {WWW}}}. \bibinfo{pages}{3248--3258}.
\newblock


\bibitem[\protect\citeauthoryear{Fu, Hua, Xie, Fang, Zhang, Sancak, Wu, Malevich, He, and Long}{Fu et~al\mbox{.}}{2024}]%
        {vcr}
\bibfield{author}{\bibinfo{person}{Dongqi Fu}, \bibinfo{person}{Zhigang Hua}, \bibinfo{person}{Yan Xie}, \bibinfo{person}{Jin Fang}, \bibinfo{person}{Si Zhang}, \bibinfo{person}{Kaan Sancak}, \bibinfo{person}{Hao Wu}, \bibinfo{person}{Andrey Malevich}, \bibinfo{person}{Jingrui He}, {and} \bibinfo{person}{Bo Long}.} \bibinfo{year}{2024}\natexlab{}.
\newblock \showarticletitle{VCR-Graphormer: {A} Mini-batch Graph Transformer via Virtual Connections}. In \bibinfo{booktitle}{\emph{Proceedings of the Twelfth International Conference on Learning Representations, {ICLR}}}.
\newblock


\bibitem[\protect\citeauthoryear{Gilmer, Schoenholz, Riley, Vinyals, and Dahl}{Gilmer et~al\mbox{.}}{2017}]%
        {message}
\bibfield{author}{\bibinfo{person}{Justin Gilmer}, \bibinfo{person}{Samuel~S. Schoenholz}, \bibinfo{person}{Patrick~F. Riley}, \bibinfo{person}{Oriol Vinyals}, {and} \bibinfo{person}{George~E. Dahl}.} \bibinfo{year}{2017}\natexlab{}.
\newblock \showarticletitle{Neural Message Passing for Quantum Chemistry}. In \bibinfo{booktitle}{\emph{International Conference on Machine Learning, {ICML}}}. \bibinfo{pages}{1263--1272}.
\newblock


\bibitem[\protect\citeauthoryear{Hamilton, Ying, and Leskovec}{Hamilton et~al\mbox{.}}{2017}]%
        {graphsage}
\bibfield{author}{\bibinfo{person}{Will Hamilton}, \bibinfo{person}{Zhitao Ying}, {and} \bibinfo{person}{Jure Leskovec}.} \bibinfo{year}{2017}\natexlab{}.
\newblock \showarticletitle{{Inductive Representation Learning on Large Graphs}}. In \bibinfo{booktitle}{\emph{Proceedings of the Advances in Neural Information Processing Systems, {NeurIPS}}}. \bibinfo{pages}{1024--1034}.
\newblock


\bibitem[\protect\citeauthoryear{He, Chen, Xu, and He}{He et~al\mbox{.}}{2022}]%
        {rlp}
\bibfield{author}{\bibinfo{person}{Qiuting He}, \bibinfo{person}{Jinsong Chen}, \bibinfo{person}{Hao Xu}, {and} \bibinfo{person}{Kun He}.} \bibinfo{year}{2022}\natexlab{}.
\newblock \showarticletitle{{Structural Robust Label Propagation on Homogeneous Graphs}}. In \bibinfo{booktitle}{\emph{{IEEE} International Conference on Data Mining, {ICDM}}}. \bibinfo{pages}{181--190}.
\newblock


\bibitem[\protect\citeauthoryear{Ji, Meng, and Zhang}{Ji et~al\mbox{.}}{2022a}]%
        {rs1}
\bibfield{author}{\bibinfo{person}{Weiyu Ji}, \bibinfo{person}{Xiangwu Meng}, {and} \bibinfo{person}{Yujie Zhang}.} \bibinfo{year}{2022}\natexlab{a}.
\newblock \showarticletitle{{SPATM:} {A} Social Period-Aware Topic Model for Personalized Venue Recommendation}.
\newblock \bibinfo{journal}{\emph{IEEE Transactions on Knowledge and Data Engineering, {TKDE}}} \bibinfo{volume}{34}, \bibinfo{number}{8} (\bibinfo{year}{2022}), \bibinfo{pages}{3997--4010}.
\newblock


\bibitem[\protect\citeauthoryear{Ji, Meng, and Zhang}{Ji et~al\mbox{.}}{2022b}]%
        {rs2}
\bibfield{author}{\bibinfo{person}{Weiyu Ji}, \bibinfo{person}{Xiangwu Meng}, {and} \bibinfo{person}{Yujie Zhang}.} \bibinfo{year}{2022}\natexlab{b}.
\newblock \showarticletitle{STARec: Adaptive Learning with Spatiotemporal and Activity Influence for {POI} Recommendation}.
\newblock \bibinfo{journal}{\emph{{ACM} Transactions on Information Systems}} \bibinfo{volume}{40}, \bibinfo{number}{4} (\bibinfo{year}{2022}), \bibinfo{pages}{65:1--65:40}.
\newblock


\bibitem[\protect\citeauthoryear{Jin, Derr, Wang, Ma, Liu, and Tang}{Jin et~al\mbox{.}}{2021}]%
        {simpGNN}
\bibfield{author}{\bibinfo{person}{Wei Jin}, \bibinfo{person}{Tyler Derr}, \bibinfo{person}{Yiqi Wang}, \bibinfo{person}{Yao Ma}, \bibinfo{person}{Zitao Liu}, {and} \bibinfo{person}{Jiliang Tang}.} \bibinfo{year}{2021}\natexlab{}.
\newblock \showarticletitle{{Node Similarity Preserving Graph Convolutional Networks}}. In \bibinfo{booktitle}{\emph{Proceedings of the ACM International Conference on Web Search and Data Mining, {WSDM}}}. \bibinfo{pages}{148--156}.
\newblock


\bibitem[\protect\citeauthoryear{Kipf and Welling}{Kipf and Welling}{2017}]%
        {gcn}
\bibfield{author}{\bibinfo{person}{Thomas~N Kipf} {and} \bibinfo{person}{Max Welling}.} \bibinfo{year}{2017}\natexlab{}.
\newblock \showarticletitle{{Semi-supervised Classification with Graph Convolutional Networks}}. In \bibinfo{booktitle}{\emph{Proceedings of the International Conference on Learning Representations, {ICLR}}}.
\newblock


\bibitem[\protect\citeauthoryear{Klicpera, Bojchevski, and G{\"{u}}nnemann}{Klicpera et~al\mbox{.}}{2019}]%
        {appnp}
\bibfield{author}{\bibinfo{person}{Johannes Klicpera}, \bibinfo{person}{Aleksandar Bojchevski}, {and} \bibinfo{person}{Stephan G{\"{u}}nnemann}.} \bibinfo{year}{2019}\natexlab{}.
\newblock \showarticletitle{{Predict then Propagate: Graph Neural Networks meet Personalized PageRank}}. In \bibinfo{booktitle}{\emph{Proceedings of the International Conference on Learning Representations, {ICLR}}}.
\newblock


\bibitem[\protect\citeauthoryear{Kohn, Hoffmann, and Scherp}{Kohn et~al\mbox{.}}{2024}]%
        {esmlp}
\bibfield{author}{\bibinfo{person}{Matthias Kohn}, \bibinfo{person}{Marcel Hoffmann}, {and} \bibinfo{person}{Ansgar Scherp}.} \bibinfo{year}{2024}\natexlab{}.
\newblock \showarticletitle{Edge-Splitting MLP: Node Classification on Homophilic and Heterophilic Graphs without Message Passing}. In \bibinfo{booktitle}{\emph{Proceedings of Learning on Graphs, {LoG}}}. \bibinfo{pages}{1--21}.
\newblock


\bibitem[\protect\citeauthoryear{Liu, Zhan, Ma, Ding, Tao, Wu, and Hu}{Liu et~al\mbox{.}}{2023}]%
        {gapformer}
\bibfield{author}{\bibinfo{person}{Chuang Liu}, \bibinfo{person}{Yibing Zhan}, \bibinfo{person}{Xueqi Ma}, \bibinfo{person}{Liang Ding}, \bibinfo{person}{Dapeng Tao}, \bibinfo{person}{Jia Wu}, {and} \bibinfo{person}{Wenbin Hu}.} \bibinfo{year}{2023}\natexlab{}.
\newblock \showarticletitle{Gapformer: Graph Transformer with Graph Pooling for Node Classification}. In \bibinfo{booktitle}{\emph{Proceedings of the International Joint Conference on Artificial Intelligence, {IJCAI}}}. \bibinfo{pages}{2196--2205}.
\newblock


\bibitem[\protect\citeauthoryear{Loshchilov and Hutter}{Loshchilov and Hutter}{2019}]%
        {adamw}
\bibfield{author}{\bibinfo{person}{Ilya Loshchilov} {and} \bibinfo{person}{Frank Hutter}.} \bibinfo{year}{2019}\natexlab{}.
\newblock \showarticletitle{{Decoupled Weight Decay Regularization}}. In \bibinfo{booktitle}{\emph{Proceedings of the International Conference on Learning Representations, {ICLR}}}.
\newblock


\bibitem[\protect\citeauthoryear{Ma, He, and Wei}{Ma et~al\mbox{.}}{2024}]%
        {polyformer}
\bibfield{author}{\bibinfo{person}{Jiahong Ma}, \bibinfo{person}{Mingguo He}, {and} \bibinfo{person}{Zhewei Wei}.} \bibinfo{year}{2024}\natexlab{}.
\newblock \showarticletitle{PolyFormer: Scalable Node-wise Filters via Polynomial Graph Transformer}. In \bibinfo{booktitle}{\emph{Proceedings of the 30th {ACM} {SIGKDD} Conference on Knowledge Discovery and Data Mining}}, \bibfield{editor}{\bibinfo{person}{Ricardo Baeza{-}Yates} {and} \bibinfo{person}{Francesco Bonchi}} (Eds.). \bibinfo{pages}{2118--2129}.
\newblock


\bibitem[\protect\citeauthoryear{Pei, Wei, Chang, Lei, and Yang}{Pei et~al\mbox{.}}{2020}]%
        {geomgcn}
\bibfield{author}{\bibinfo{person}{Hongbin Pei}, \bibinfo{person}{Bingzhe Wei}, \bibinfo{person}{Kevin~Chen{-}Chuan Chang}, \bibinfo{person}{Yu Lei}, {and} \bibinfo{person}{Bo Yang}.} \bibinfo{year}{2020}\natexlab{}.
\newblock \showarticletitle{Geom-GCN: Geometric Graph Convolutional Networks}. In \bibinfo{booktitle}{\emph{Proceedings of the International Conference on Learning Representations, {ICLR}}}.
\newblock


\bibitem[\protect\citeauthoryear{Ramp{\'{a}}sek, Galkin, Dwivedi, Luu, Wolf, and Beaini}{Ramp{\'{a}}sek et~al\mbox{.}}{2022}]%
        {gps}
\bibfield{author}{\bibinfo{person}{Ladislav Ramp{\'{a}}sek}, \bibinfo{person}{Mikhail Galkin}, \bibinfo{person}{Vijay~Prakash Dwivedi}, \bibinfo{person}{Anh~Tuan Luu}, \bibinfo{person}{Guy Wolf}, {and} \bibinfo{person}{Dominique Beaini}.} \bibinfo{year}{2022}\natexlab{}.
\newblock \showarticletitle{{Recipe for a General, Powerful, Scalable Graph Transformer}}. In \bibinfo{booktitle}{\emph{Proceedings of the Advances in Neural Information Processing Systems, {NeurIPS}}}, Vol.~\bibinfo{volume}{35}. \bibinfo{pages}{14501--14515}.
\newblock


\bibitem[\protect\citeauthoryear{Veli{\v{c}}kovi{\'c}, Cucurull, Casanova, Romero, Lio, and Bengio}{Veli{\v{c}}kovi{\'c} et~al\mbox{.}}{2018}]%
        {gat}
\bibfield{author}{\bibinfo{person}{Petar Veli{\v{c}}kovi{\'c}}, \bibinfo{person}{Guillem Cucurull}, \bibinfo{person}{Arantxa Casanova}, \bibinfo{person}{Adriana Romero}, \bibinfo{person}{Pietro Lio}, {and} \bibinfo{person}{Yoshua Bengio}.} \bibinfo{year}{2018}\natexlab{}.
\newblock \showarticletitle{{Graph Attention Networks}}. In \bibinfo{booktitle}{\emph{Proceedings of the International Conference on Learning Representations, {ICLR}}}.
\newblock


\bibitem[\protect\citeauthoryear{Wang, Zhu, Bo, Cui, Shi, and Pei}{Wang et~al\mbox{.}}{2020}]%
        {amgcn}
\bibfield{author}{\bibinfo{person}{Xiao Wang}, \bibinfo{person}{Meiqi Zhu}, \bibinfo{person}{Deyu Bo}, \bibinfo{person}{Peng Cui}, \bibinfo{person}{Chuan Shi}, {and} \bibinfo{person}{Jian Pei}.} \bibinfo{year}{2020}\natexlab{}.
\newblock \showarticletitle{{AM-GCN:} Adaptive Multi-channel Graph Convolutional Networks}. In \bibinfo{booktitle}{\emph{Proceedings of the {ACM} {SIGKDD} Conference on Knowledge Discovery and Data Mining}}. \bibinfo{pages}{1243--1253}.
\newblock


\bibitem[\protect\citeauthoryear{Wang, Zhang, Zhang, and Ye}{Wang et~al\mbox{.}}{2025}]%
        {simmlp}
\bibfield{author}{\bibinfo{person}{Zehong Wang}, \bibinfo{person}{Zheyuan Zhang}, \bibinfo{person}{Chuxu Zhang}, {and} \bibinfo{person}{Yanfang Ye}.} \bibinfo{year}{2025}\natexlab{}.
\newblock \showarticletitle{Training mlps on graphs without supervision}. In \bibinfo{booktitle}{\emph{Proceedings of the ACM International Conference on Web Search and Data Mining, {WSDM}}}. \bibinfo{pages}{697--706}.
\newblock


\bibitem[\protect\citeauthoryear{Wu, Souza, Zhang, Fifty, Yu, and Weinberger}{Wu et~al\mbox{.}}{2019}]%
        {sgc}
\bibfield{author}{\bibinfo{person}{Felix Wu}, \bibinfo{person}{Amauri Souza}, \bibinfo{person}{Tianyi Zhang}, \bibinfo{person}{Christopher Fifty}, \bibinfo{person}{Tao Yu}, {and} \bibinfo{person}{Kilian Weinberger}.} \bibinfo{year}{2019}\natexlab{}.
\newblock \showarticletitle{{Simplifying Graph Convolutional Networks}}. In \bibinfo{booktitle}{\emph{Proceedings of the International Conference on Machine Learning, {ICML}}}. \bibinfo{pages}{6861--6871}.
\newblock


\bibitem[\protect\citeauthoryear{Wu, Zhao, Li, Wipf, and Yan}{Wu et~al\mbox{.}}{2022}]%
        {nodeformer}
\bibfield{author}{\bibinfo{person}{Qitian Wu}, \bibinfo{person}{Wentao Zhao}, \bibinfo{person}{Zenan Li}, \bibinfo{person}{David Wipf}, {and} \bibinfo{person}{Junchi Yan}.} \bibinfo{year}{2022}\natexlab{}.
\newblock \showarticletitle{{NodeFormer: A Scalable Graph Structure Learning Transformer for Node Classification}}. In \bibinfo{booktitle}{\emph{Proceedings of the Advances in Neural Information Processing Systems, {NeurIPS}}}, Vol.~\bibinfo{volume}{35}. \bibinfo{pages}{27387--27401}.
\newblock


\bibitem[\protect\citeauthoryear{Wu, Zhao, Yang, Zhang, Nie, Jiang, Bian, and Yan}{Wu et~al\mbox{.}}{2023}]%
        {sgformer}
\bibfield{author}{\bibinfo{person}{Qitian Wu}, \bibinfo{person}{Wentao Zhao}, \bibinfo{person}{Chenxiao Yang}, \bibinfo{person}{Hengrui Zhang}, \bibinfo{person}{Fan Nie}, \bibinfo{person}{Haitian Jiang}, \bibinfo{person}{Yatao Bian}, {and} \bibinfo{person}{Junchi Yan}.} \bibinfo{year}{2023}\natexlab{}.
\newblock \showarticletitle{Simplifying and empowering transformers for large-graph representations}. In \bibinfo{booktitle}{\emph{Proceedings of the Advances in Neural Information Processing Systems, {NeurIPS}}}.
\newblock


\bibitem[\protect\citeauthoryear{Xing, Wang, Li, Huang, and Shi}{Xing et~al\mbox{.}}{2024}]%
        {cob}
\bibfield{author}{\bibinfo{person}{Yujie Xing}, \bibinfo{person}{Xiao Wang}, \bibinfo{person}{Yibo Li}, \bibinfo{person}{Hai Huang}, {and} \bibinfo{person}{Chuan Shi}.} \bibinfo{year}{2024}\natexlab{}.
\newblock \showarticletitle{Less is More: on the Over-Globalizing Problem in Graph Transformers}. In \bibinfo{booktitle}{\emph{Proceedings of the International Conference on Machine Learning, {ICML}}}.
\newblock


\bibitem[\protect\citeauthoryear{Xu, Wang, Wu, Wen, Zhao, and Wan}{Xu et~al\mbox{.}}{2024}]%
        {Fraud_Detection_1}
\bibfield{author}{\bibinfo{person}{Fan Xu}, \bibinfo{person}{Nan Wang}, \bibinfo{person}{Hao Wu}, \bibinfo{person}{Xuezhi Wen}, \bibinfo{person}{Xibin Zhao}, {and} \bibinfo{person}{Hai Wan}.} \bibinfo{year}{2024}\natexlab{}.
\newblock \showarticletitle{Revisiting Graph-Based Fraud Detection in Sight of Heterophily and Spectrum}. In \bibinfo{booktitle}{\emph{Thirty-Eighth {AAAI} Conference on Artificial Intelligence, {AAAI}}}. \bibinfo{pages}{9214--9222}.
\newblock


\bibitem[\protect\citeauthoryear{Xu, Li, Tian, Sonobe, Kawarabayashi, and Jegelka}{Xu et~al\mbox{.}}{2018}]%
        {jknet}
\bibfield{author}{\bibinfo{person}{Keyulu Xu}, \bibinfo{person}{Chengtao Li}, \bibinfo{person}{Yonglong Tian}, \bibinfo{person}{Tomohiro Sonobe}, \bibinfo{person}{Ken{-}ichi Kawarabayashi}, {and} \bibinfo{person}{Stefanie Jegelka}.} \bibinfo{year}{2018}\natexlab{}.
\newblock \showarticletitle{Representation Learning on Graphs with Jumping Knowledge Networks}. In \bibinfo{booktitle}{\emph{Proceedings of the 35th International Conference on Machine Learning}} \emph{(\bibinfo{series}{Proceedings of Machine Learning Research}, Vol.~\bibinfo{volume}{80})}. \bibinfo{pages}{5449--5458}.
\newblock


\bibitem[\protect\citeauthoryear{Ying, Cai, Luo, Zheng, Ke, He, Shen, and Liu}{Ying et~al\mbox{.}}{2021}]%
        {graphormer}
\bibfield{author}{\bibinfo{person}{Chengxuan Ying}, \bibinfo{person}{Tianle Cai}, \bibinfo{person}{Shengjie Luo}, \bibinfo{person}{Shuxin Zheng}, \bibinfo{person}{Guolin Ke}, \bibinfo{person}{Di He}, \bibinfo{person}{Yanming Shen}, {and} \bibinfo{person}{Tie-Yan Liu}.} \bibinfo{year}{2021}\natexlab{}.
\newblock \showarticletitle{{Do Transformers Really Perform Badly for Graph Representation}}. In \bibinfo{booktitle}{\emph{Proceedings of the Advances in Neural Information Processing Systems, {NeurIPS}}}. \bibinfo{pages}{28877--28888}.
\newblock


\bibitem[\protect\citeauthoryear{Zeng, Zhou, Srivastava, Kannan, and Prasanna}{Zeng et~al\mbox{.}}{2020}]%
        {GraphSAINT}
\bibfield{author}{\bibinfo{person}{Hanqing Zeng}, \bibinfo{person}{Hongkuan Zhou}, \bibinfo{person}{Ajitesh Srivastava}, \bibinfo{person}{Rajgopal Kannan}, {and} \bibinfo{person}{Viktor~K. Prasanna}.} \bibinfo{year}{2020}\natexlab{}.
\newblock \showarticletitle{{GraphSAINT: Graph Sampling Based Inductive Learning Method}}. In \bibinfo{booktitle}{\emph{Proceedings of the International Conference on Learning Representations, {ICLR}}}.
\newblock


\bibitem[\protect\citeauthoryear{Zhang, Liu, Hu, and Lee}{Zhang et~al\mbox{.}}{2022}]%
        {ansgt}
\bibfield{author}{\bibinfo{person}{Zaixi Zhang}, \bibinfo{person}{Qi Liu}, \bibinfo{person}{Qingyong Hu}, {and} \bibinfo{person}{Chee{-}Kong Lee}.} \bibinfo{year}{2022}\natexlab{}.
\newblock \showarticletitle{{Hierarchical Graph Transformer with Adaptive Node Sampling}}. In \bibinfo{booktitle}{\emph{Proceedings of the Advances in Neural Information Processing Systems, {NeurIPS}}}. \bibinfo{pages}{21171--21183}.
\newblock


\bibitem[\protect\citeauthoryear{Zhao, Li, Wen, Wang, Liu, Sun, Xie, and Ye}{Zhao et~al\mbox{.}}{2021}]%
        {gophormer}
\bibfield{author}{\bibinfo{person}{Jianan Zhao}, \bibinfo{person}{Chaozhuo Li}, \bibinfo{person}{Qianlong Wen}, \bibinfo{person}{Yiqi Wang}, \bibinfo{person}{Yuming Liu}, \bibinfo{person}{Hao Sun}, \bibinfo{person}{Xing Xie}, {and} \bibinfo{person}{Yanfang Ye}.} \bibinfo{year}{2021}\natexlab{}.
\newblock \showarticletitle{{Gophormer: Ego-Graph Transformer for Node Classification}}.
\newblock \bibinfo{journal}{\emph{arXiv preprint arXiv:2110.13094}} (\bibinfo{year}{2021}).
\newblock


\bibitem[\protect\citeauthoryear{Zhuo, Liu, Lu, Ma, Fu, Wang, Guo, Wang, Cao, and Yang}{Zhuo et~al\mbox{.}}{2025}]%
        {dual}
\bibfield{author}{\bibinfo{person}{Jiaming Zhuo}, \bibinfo{person}{Yuwei Liu}, \bibinfo{person}{Yintong Lu}, \bibinfo{person}{Ziyi Ma}, \bibinfo{person}{Kun Fu}, \bibinfo{person}{Chuan Wang}, \bibinfo{person}{Yuanfang Guo}, \bibinfo{person}{Zhen Wang}, \bibinfo{person}{Xiaochun Cao}, {and} \bibinfo{person}{Liang Yang}.} \bibinfo{year}{2025}\natexlab{}.
\newblock \showarticletitle{{DUALFormer: Dual Graph Transformer}}. In \bibinfo{booktitle}{\emph{Proceedings of the International Conference on Learning Representations}}.
\newblock


\end{thebibliography}
